\documentclass[a4paper,11pt]{article}
\pdfoutput=1
\usepackage{jmlr2e}
\usepackage[dvipsnames]{xcolor}
\usepackage[a4paper,lmargin=1.25in,rmargin=1.25in,bottom=1.25in,top=1in]{geometry}
\usepackage{tikz-cd}
\usepackage{colortbl}
\usepackage{xspace}
\usepackage{microtype}
\usepackage{formatmacros}
\usepackage{stmaryrd}

\usepackage{diffcoeff}
\usepackage[T1]{fontenc}
\usepackage{mlmodern}  
\usepackage[natbib=true,block=space,style=authoryear-comp,maxbibnames=9,uniquename=false,sortcites=true]{biblatex}   
\usepackage{notation}

\def\expto{\mathrel{\rightsquigarrow}}

\NewDocumentCommand\varfrac{s m m}{
    \def\sfrac {\raisebox{.5ex}{\(\scriptstyle#2\)}
                \mkern-3mu/\mkern-3mu
                \raisebox{-.5ex}{\(\scriptstyle#3\)}}
    \def\ssfrac{\raisebox{.3ex}{\(\scriptscriptstyle#2\)}
                \scriptstyle{\mkern-3mu/\mkern-3mu}
                \raisebox{-.3ex}{\(\scriptscriptstyle#3\)}}
    \mathinner{
        \IfBooleanTF#1{
            \mathchoice{\sfrac}{\frac{#2}{#3}}{\frac{#2}{#3}}{\frac{#2}{#3}}
        }{
            \mathchoice{\frac{#2}{#3}}{\sfrac}{\ssfrac}{\ssfrac}
        }
    }
}

\newcommand{\Tfrac}[2]{%
  \ooalign{%
    $\genfrac{}{}{1.2pt}1{#1}{#2}$\cr%
    $\color{white}\genfrac{}{}{.4pt}1{\phantom{#1}}{\phantom{#2}}$}%
}
\allowdisplaybreaks  

\title{Information Processing Equalities\\
       and the Information--Risk Bridge}

\jmlrheading{?}{2022}{$n$--$(n+l-1)$}{9/22}{?}{Williamson, Cranko} 
\ShortHeadings{Information Processing}{Williamson, Cranko}
 \author{%
    Robert C. Williamson \email{Bob.Williamson@uni-tuebingen.de}\\
    \addr{University of T\"{u}bingen and T\"{u}bingen AI Center,\\ Germany} 
    \AND
    Zac Cranko \email{Zac.Cranko@gmail.com}\\
    \addr{Sydney, Australia }
}
\editor{?}
\begin{document}

\maketitle
\begin{abstract}
  We introduce two new classes of measures of information for statistical
  experiments which generalise and subsume $\phi$-divergences, integral
  probability metrics,  $\mathfrak{N}$-distances (MMD), and $(f,\Gamma)$
  divergences between  two or more distributions.  This enables us to
  derive a simple geometrical relationship between measures of information
  and the Bayes risk of a statistical decision problem, thus extending the
  variational $\phi$-divergence representation to multiple distributions in
  an entirely symmetric manner.  The new families of divergence are closed
  under the action of Markov operators which yields an information
  processing \emph{equality} which is a refinement  and generalisation of
  the classical data processing \emph{inequality}. This equality gives
  insight into the significance of the choice of the hypothesis class in
  classical risk minimization.
\end{abstract}


\section{Introduction}
\label{sec:introduction}
\begin{flushright}
\vspace*{-2mm}
  \begin{minipage}{0.5\textwidth}
    \begin{flushright}
      \footnotesize
      \textit{
        A key word in statistics is information\ldots But what is information?
        No other concept in statistics is more elusive in its meaning and less amenable to a generally agreed definition.} ---~Debabrata \textcite[p.~1]{Basu:1975aa}.
    \end{flushright}
  \end{minipage}
\end{flushright}
Machine learning is information processing.  But what ``information'' is
meant?  Choosing exactly how to
measure information has become topical of late in machine learning, with
methods such as GANs predicated on the notion of being unable to compute a
likelihood function, but being able to measure an information distance
between a target and synthesised distribution \autocite{Binkowski:2018aa}.
Commonly used measures include the the Shannon information/entropy of a
single distribution and the Kullback-Leibler divergence or Variational
divergence between two different distributions.  \citeauthor{Csiszar1967}'s
$\phi$-entropies and $\phi$-divergences \autocite{csiszar63,Csiszar1967}
subsume these and many other divergences, and satisfy the famous
information processing inequality \autocite{Ziv1973} which states that the
amount of information can only decrease (or stay constant) as a result of
``information processing.''

The present paper presents a new and general definition of information that
subsumes many in the literature.  The key novelty of the paper is
the redefinition of classical measures of information as expected values of the
support function of particular convex sets. The advantage of this
redefinition is that it provides a surprising insight into the classical
information processing inequality, which can consequently be seen to be an
\emph{equality} albeit one with different measures of information on either
side of the equality. The reformulation also enables an elegant proof of
the 1:1 relationship between information and (Bayes) risk, showing in an
unambiguous way that there can not be a sensible definition of information
that does not take account of the use to which the information will be put.

The rest of the paper is organised as follows. In the remainder of the
present section, we introduce the $\phi$-divergence, summarise earlier work
on extending it to several distributions, and sketch a philosophy of
information which our main theoretical results formally justify and
support.  In \S\ref{sec:technical-background} we present the necessary
technical tools we use; \S\ref{sec:information-defs} presents  the general
``unconstrained'' information measures (with no restriction on the model
class); \S\ref{sec:bridge} presents the bridge between information measures
and the (unconstrained) Bayes risk; \S\ref{sec:Fscr-Nfrak-info} 
presents the constrained measures of
information (where there is a restriction on the model class), as well as
the generalisation of the ``bridge'' to this case; \S\ref{sec:conclusion}
concludes. There are
four appendices: Appendix \ref{sec:var-rep} relates our definition of
$D$-information to the classical  variational representation of a (binary)
$\phi$-divergence. Appendix  \ref{sec:expected-gauge} shows how our
measure of information is naturally viewed as an expected gauge function.
Appendix
\ref{sec:entropies} examines the different entropies induced by the
$\mathfrak{F}$-information, showing how they too implicitly have a model
class hidden inside their definition.
Finally, Appendix \ref{sec:precursors} summarises earlier attempts to generalise
$\phi$-divergences to take account of a model class\footnote{Part 
	II of the present paper \citep{Williamson:2023aa}, to appear 
	in due course, 
	will contain an explanation of the relationship
	between our information processing theorems and the traditional
	inequalities (usually couched in terms of mutual information); the
	derivation of classical data processing theorems for divergences (with the
	same measure of information on either side of the inequality) from the
	results in part I; relationships to measures of informativity of
	observation channels; and relationships to existing results connecting
	information and estimation theory.}.

\subsection{The \texorpdfstring{$\phi$}{phi}-divergence}
Suppose $\mu,\nu$ are two probability distributions,  with $\mu$ absolutely
continuous with respect to $\nu$ and let
\begin{gather}
  \Def{\Phi}{\left\{\phi\st\R++\to\R,\quad\text{$\phi$ convex},  \quad \phi(1) = 0\right\}}.
  \label{eq:phi_desiderata}
\end{gather}
For $\phi\in\Phi$, the \Def{$\phi$-divergence} between $\mu$ and $\nu$ is defined as
\begin{gather}
	\Def{\I_\phi(\mu,\nu)}{\int \phi\rbr{\diff{\mu}{\nu}}\dv \nu}.
  \label{def:fdiv_fmi}
\end{gather}
Popular examples of $\phi$-divergences include the Kullback-Liebler
divergence ($\phi(t)=t\log t$) and the Variational divergence
($\phi(t)=|t-1|$) among others;  see \citep{Reid2011}. 

There are two existing classes of extensions to binary $\phi$-divergences
--- devising measures of information for more than two distributions, and
restricting the implicit optimization in the variational form (see Appendix
\ref{sec:var-rep}) as a form of regularisation.  We summarise work along
the first of these lines in the next subsection, and the second in
\autoref{sec:precursors} after we have introduced the necessary concepts to
make sense of these attempts.

\subsection{Beyond Binary --- ``\texorpdfstring{$\phi$}{phi}-divergences'' 
for more than two distributions}
Earlier attempts to extend $\phi$-divergences beyond the case of two distributions 
include the \Def{$\phi$-affinity} between $n>2$ distinct distributions; this is
also known as the Matusita affinity \citep{matusita67,Matusita1971}, the
\Def{$f$-dissimilarity} \citep{gyorfi1975f, gyorfi1978f},  the generalised
$\phi$-divergence \citep{Ginebra:2007} or (on which we build in the present
paper) \Def{$D$-divergences}~\citep{Gushchin:2008aa}.  One could conceive of
these as ``$n$-way distances'' \citep{Warrens2010} but most of the
intuition about distances does not carry across, and so we will not adopt
such an interpretation, and in the body of the paper refer to the objects
simply as ``measures of information.''
 
Generalisations of \emph{particular} divergences to several distributions
include the \Def{information radius} \citep{Sibson1969}
$R(P_1,\ldots,P_k)=\frac{1}{k}\sum_{i=1}^k
\mathrm{KL}\left(P_i,(P_1+P_2+\cdots+P_k)/k\right)$ where
$\mathrm{KL}(P,Q)$ is the Kullback-Leibler divergence and the \Def{average
divergence} \citep{Sgarro1981}
$K(P_1,\ldots,P_k)=\frac{1}{k(k-1)}\sum_{i=1}^k\sum_{j=1}^k
\mathrm{KL}(P_i,P_j)$.  Some other approaches to generalising
$\phi$-divergences to more than two distributions are summarised by
\cite{Basseville2010}.

The general multi-distribution divergence has been used in hypothesis
testing \citep{Menendez2005,Zografos1998}. \citet{gyorfi1975f} bounded the
minimal probability of error in terms of the $f$-affinity; see also
\citep{Glick1973,toussaint78}.  These results are analogous to surrogate
regret bounds \citep[section 7.1]{Reid2011} because there is 
in fact an \emph{exact} relationship between
$\I_\phi$ and the Bayes risk of an associated multiclass classification
problem; see  \S\ref{sec:bridge}.  
Multidistribution $\phi$-divergences have also been used  to
extend rate-distortion theory (primarily as a technical means to get better
bounds) \citep{Zakai1975} and to unify information theory with  the second
law of thermodynamics \citep{Merhav2011}.  The \emph{estimation} of these
divergences has been studied by \citet{Morales1998}. The
connection to Bayes risk suggests alternate estimation schemes.  

Going in the opposite direction, it is worth noting that the entropy of a
\emph{single} distribution can be viewed as the $\phi$-divergence between
the given distribution and a reference (or ``uniform'') distribution
\citep{Torgersen:1981}; see also Appendix \ref{sec:entropies}.
\subsection{Information is as Information Does}
In developing a philosophy of information, \citet[page 20]{Adriaans:2008wg}
adopted  the slogan ``No information without
transformation!''   They asked ``what does information \emph{do} for each
process?''  We reverse this to: ``what does each process
\emph{do} to information?''

We avoid an essentialist claim of ``one true notion'' of information, but
do not feel it necessary to follow the example of 
\citet{Csiszar1972} of eschewing the word ``information'' for the 
neologism ``informativity.'' We believe that the elements of our field need
to prove their mettle by their relationships. Barry \citet*[section
3]{Mazur:2008aa} observed that ``mathematical objects [are] determined
by the network of relationships they enjoy with all the other objects of
their species''  and proposed to ``subjugate the role of the mathematical
object to the role of its network of relationships --- or, a further
extreme --- simply replace the mathematical object by this
network''.\footnote{This perspective is sometimes described as 
	``Grothendieck's relative point of view'' in mathematics, but the
	insight holds more generally: ``We only understand something
	according to the transformations that can be performed on it'' 
	--- Michel \citet*[page 9]{serres1974hermes}, 
	quoted in \citep[page 41]{sack2019software}.
	}

One could argue that such systematic study of the elements and their
fundamental transformations is essential to achieve the called for
transition of machine learning from alchemy to a mature science
\citep{Rahimi:2017aa}.  We make a small step in this direction, focusing
upon the transformation that measures of information of an experiment
undergo when the experiment is observed via a noisy observation channel.
This is a return to roots, since the very notion of Shannon information
information was motivated by communication over noisy channels
\citep{Shannon1948,Shannon:1949aa}, and that of the Kullback-Leibler
divergence motivated by notions of sufficiency \citep{Kullback1951}. That a
sufficient statistic can be viewed as the output of a noisy observation
channel is made precise in  the general definition 
of sufficiency and approximate sufficiency due to 
\citet{LeCam:1964wk}.

Our perspective is motivated by the largely forgotten conclusion of
\citet{DeGroot1962}, that even if one is only seeking some vague sense of
``information'' in data, ultimately one will \emph{use} this
``information'' through some act (else why bother?), and such acts incur a
utility (or loss), which can be quantified\footnote{Interestingly, DeGroot
	was motivated to extend the attempt of \citet{Lindley1956}  
	to quantify the ``amount of
	information'' in an experiment, but unlike Lindley, 
	did not presume that this was necessarily Shannon information.}. 
Thus any \emph{use}ful notion
of information needs to take account of utility.  Our general notion of
information of an experiment is consistent with DeGroot's utilitarian
premise; we suggest that it is the most general such concept
consistent with the precepts of decision theory and statistical learning
theory. 

This philosophy is made precise by our results showing the 
equivalence of the measures of information (which subsume most 
of those in the literature) and the Bayes risk of 
a statistical decision problem. Significantly, this means that \emph{the 
choice of a measure of information is equivalent to the choice 
of a loss function (plus, potentially, the choice of a convex model class)}
--- thus any notion of information subsumed by our
general measures really encodes the use to which one envisages 
the information being put, as De Groot admonished 60 years ago.

\section{Technical Background and Notation}
\label{sec:technical-background}


For positive integer $n$,  we write $[n]\defeq \{1,2,\ldots,n\}$.
Let $\Def{\e_i}$ denote the $i$-th canonical unit vector, and
$\Def{1_n}{(1,\ldots,1)\in\R^n}$. 
We use standard concepts of convex analysis\footnote{
	See
	\citep{hiriarturruty2001fca,Rockafellar:1970,Penot2012Calculus,
		Bauschke2011Convex, aliprantis2006infinite}.
	Since notation in the literature varies, we spell out our choice
	in full.}.
Let $\Def{\Rx}{[-\infty,\infty]}$ and $\Def{\Rx_+}{[0,\infty]}$.
Let $f\colon X\to\Rx$. 
Its \Def{domain} $\Def{\dom f} \defeq \{x\in X\colon f(x)\ne \infty\}$ and
its \Def{Legendre-Fenchel conjugate}, 
\[
	\Def{f^*(x^*)}{\sup_{x\in X}\rbr{\inp{x^*,x} - f(x)}}.
\]
If $f$ is proper, closed, and convex, it is equal to its
biconjugate: $f= (f^*)^*$. The \Def{epigraph} and
\Def{hypograph} of $f$ are the sets 
\[
	\Def{\epi f}{\cbr{(x,t) \in \dom(f)\times\R\st t\geq f(x)}} 
	\mbox{\  \&\  }
	\Def{\hyp f}{\cbr{(x,t) \in \dom(f)\times\R\st t\leq f(x)}}.
\]
The function $f$ is \Def{closed and convex} if and only
if the set $\epi(f)$ (or equivalently $\hyp(-f)$) is also. The
\Def{subdifferential} of $f$ at $x\in X$ is the set 
\[
	\Def{\subdiff f(x)}{
\cbr{x^* \in X^* \st \forall{y\in\dom f} f(y) - f(x) \geq
\inp{x^*, y - x} }}.
\]
The domain of the differential is the set 
$\Def{\dom\subdiff f}{ \cbr{x\in X \st \subdiff f(x) \neq\emptyset}}$.
A \Def{selection} is a mapping $\nabla f\colon  \dom
\subdiff f \to X^*$ that satisfies $\nabla f(x) \in ∂f(x)$ for all
$x\in\dom ∂f$, and it is commonly abbreviated to  $\nabla f \in ∂f$. If
$\partial f$ is a singleton, then $\partial f$ corresponds to the
classical differential which we write $\Def{\operatorname{D} f}$.

For
$f\colon X\rightarrow \bar{\R}$ and $\alpha\in\R$, the \Def{$\alpha$
below level set of $f$} is
\[
\Def{\lev_{\le\alpha}(f)}{\cbr{x\in X\st f(x)\le \alpha}}.
\]
If $f\colon\R\rightarrow\R$ then its \Def{perspective} is the function 
$\breve{f}\colon\R\times\R\rightarrow \R$ given by 
$\Def{\breve{f}(x,y)= yf(x/y)}$.  
The perspective $\breve{f}$ is
positively homogeneous and is convex whenever $f$ is. Observe that
$\breve{f}(x,1)=f(x)$.
The \Def{halfspace} with normal $1_n$ (and zero offset) is
$\Def{\halfspace^\le_{1_n}}{\lev_{\leq 0}\inp{\marg, 1_n}}$
  
We use $\hadamard$ for the \Def{Hadamard product}: that is, if $X\ni
f,g$ is a function space then $f\hadamard g$ is the regular function
product $\Def{(f\hadamard g)(\marg)}{f(\marg)g(\marg)}$; if $X$  has
dimension $n<\infty$ then 
element-wise vector product is written
$\Def{(f\hadamard g)}{(f_1g_1, \dots, f_ng_n)}$,

For $S,T\subseteq X$ and $x\in X$,   $\Def{S+x}{ \cbr{s+x\st s\in S}}$, and  
$\Def{S+T}{ \cbr{S+t\st t\in T}}$
(the \Def{Minkowski sum}).
For $S\subseteq X$ we associate two functions: the \Def{$S$ support
function}, 
\begin{gather}
	\label{eq:support-function-def}
	\Def{\sprt_S(x^*)}{ \sup_{x\in S}\inp{x^*, x}}
\end{gather}
and the
\Def{$S$ indicator function} 
\[
	\Def{\ind_S(x)}{ \infty·\iver{x\in S}},
\]
where $\Def{\iver{p}}=1$ if $p$ is true and 0 otherwise,
and we adopt the convention that
$\infty \iver{\mathrm{true}} = 0$.
If $S$ is closed and convex then the support function is the Fenchel
conjugate of the indicator function and vice versa.  
The \Def{recession cone of\, $S$} is the set
\[
\Def{\rec S}{\cbr{d\in X \st S + d = S}}.
\]
If $S$ is convex then $\rec S$ is convex. If $X$
is finite dimensional and $S$ is bounded then $\rec S = \cbr{0}$. 
The \Def{polar cone} of $S$
is the set 
\begin{gather}
	\label{eq:polar-cone-def}
	\Def{S^*}{\cbr{x^*\in X^*\st\forall{s\in S}\!\! \inp{x^*,s} \leq 0}}.
\end{gather}The \Def{dual cone} (negative polar cone) of $S$
is the set 
\begin{gather}
	\label{eq:dual-cone-def}
	\Def{S^+}{\cbr{x^*\in X^*\st\forall{s\in S}\!\! \inp{x^*,s} \geq 0}}.
\end{gather}
The 
\Def{convex hull of $S$} is the set 
\[
\Def{\conv S}{\bigcap \cbr{T \subseteq X \st S\subseteq T,\ \mbox{$T$ is
convex}}},
\]
the \Def{closed convex hull of $S$} is the set which we abbreviate as
$\Def{\cl\conv S}{\cl(\conv S)}$.

For two measurable spaces $(X,\Sigma_X)$ and $(Y,\Sigma_Y)$ the notation
$\Def{f\colon (X,\Sigma_X) \to (Y,\Sigma_Y)}$ means that $f$ is a measurable function
with respect to the respective $\sigma$-algebras, which it is often
convenient to abbreviate to $f: X \to (Y,\Sigma_Y)$.  The \Def{Borel
$\sigma$-algebra on a set $X$} with some topology is $\Def{\borel(X)}$, and we
write $\Def{(X,\borel)}{(X,\borel(X))}$.
The set of proper, closed, convex and measurable sets $S\in \Sigma_X$ is
$\Def{\cvx(X)}$.
The subcollection of these that recess in directions at most $T\subseteq X$ is
\[
	\Def{\cvxrec(S, T)}{ \cbr{ D\in\cvx(S) \st \rec D \subseteq T}}. 
\]

Let $\Def{\probm(X)}$ be the set of probability measures on a measurable space
$(X,\Sigma_X)$. If $X$ has dimension $n<\infty$ this is isomorphic to the
set of
vectors $\cbr{p\in \R^n \st p_i \geq 0, \sum_i p_i = 1}$ and its relative interior
$\relint\probm(X)$ is the subset of vectors for which $p_i>0$ for each
$i\in[n]$. 
If $f\colon X\rightarrow\R$ and $\mu\in\Delta(X)$, we write 
$\Def{\mu f=\mu(f)\coloneqq\int f\dd\mu}$.
Conventionally a \Def{Markov kernel} is a function $M\colon  Y\times\Sigma_X \to \R$
which is $\Sigma_Y$-measurable in its first argument and a probability
measure over $X$ in its second. We use the notation of \citet{Cinlar:2011aa}
to more compactly write $\Def{M\colon Y\expto X}$. When $Y$ has dimension
$n<\infty$ we call a Markov kernel $E\colon Y\expto X$ an \Def{experiment}. 
%
%
It is convenient to
stack the distributions $E(1),\ldots,E(n)$ induced by $E$ into a vector of 
measures (one for each $y\in Y$), the notation
for which we overload: $\Def{E}{(E_1,\dots, E_n)}$. Note that while $E$ is
an experiment (Markov kernel), $E_i$ ($i\in[n]$) are measures. If $\mu$ is a measure
that dominates each $E_i$, then the vector of Radon-Nikodym derivatives
with respect to $\mu$ is 
\[
	\Def{\diff E/\mu}{\left(\diff{E_1}/\mu,\dots, \diff{E_n}/\mu\right)} 
\]
and as a function maps $X\to\R^n_{\geq 0}$.\footnote{While
this overloading may appear overeager, it provides substantial
simplification subsequently.}
An experiment
$\Def{E^{\mathrm{tni}}}\colon Y\expto X$ with
${ ({\dd E^{\mathrm{tni}}}/{ \dd\rho}) (x) }={c(x)1_n}$, for some 
$c(x)>0$ is a \Def{totally noninformative experiment}. Conversely, an
experiment $\Def{E^{\mathrm{ti}}}\colon Y \expto X$ is a
\Def{totally informative experiment} if for all $A\in \Sigma_X$, for
all $i\ne j$, $E_i^{\mathrm{ti}}(A)>0 \implies E_j^{\mathrm{ti}}(A)=0$
\citep{torgersen1991cse}.
When $X=Y=[n]$, $E\colon Y\expto X$ can be represented by an $n\times n$
stochastic matrix.

For the following definitions, fix measurable spaces $(\Omega_1,\Sigma_1)$ 
and $(\Omega_2,\Sigma_2)$. The \Def{measurable functions} $\Omega_1\expto\Omega_2$ 
are $\Def{\measf(\Omega_1,\Omega_2)}$ and 
$\Def{\measf(\Omega)\defeq\measf(\Omega,\R)}$ 
refers to the \Def{real measurable functions.} The \Def{signed measures} on $\Omega$
are $\Def{\signm(\Omega)}$, the subset of these which are \Def{probability
measures} is $\Def{\probm(\Omega)}$. To a probability 
measure $\mu\in\probm(\Omega)$ we 
associate the \Def{expectation functional}
\begin{gather}
  \begin{aligned}
	  \mu :\measf(\Omega_1)\to \R, \quad \Def{\mu f}{\int \mu(\dv x)
	  f(x)}.
  \end{aligned}
\end{gather}
There are two operators associated to and conventionally overloaded with $E$\footnote{Note the postfix notation for action of $E$ on probability measures.}:
\begin{gather}
  \begin{aligned}
    E & :\measf(\Omega_2)\to \measf(\Omega_1)
    \\\Def{Ef(x_1)} &\defeq \Def{\int_{\Omega_2} E(x_1, \dv x_2) f(x_2)},
  \end{aligned}
  \quad
  \begin{aligned}
    E & :\probm(\Omega_1)\to \probm(\Omega_2)
    \\\Def{\mu E(\dv x_2)} &\defeq \Def{\int_{\Omega_1} \mu(\dv x_1)E(x_1,
    \dv x_2)}.
  \end{aligned}
  \label{def:markov_operators}
\end{gather}
The definitions above make it convenient to 
chain experiments:
\begin{gather}
	\Def{E_1E_2(\omega_1, \dv \omega_3)} \defeq\Def{\int_{\Omega_2} 
E(\omega_1, \dv \omega_2)E(\omega_2, \dv\omega_3)},
\end{gather}
where $E_1:\Omega_1\expto\Omega_2$ and $E_2:\Omega_2\expto \Omega_3$; 
thus $E_1E_2:\Omega_1\expto \Omega_3$.

It is common in the information theory literature to write  $\mathsf{X}\to
\mathsf{Y}\to \mathsf{Z}$ to denote random variables $\mathsf{X}$,
$\mathsf{Y}$ and  $\mathsf{Z}$ which form a \Def{Markov chain}; that is,
$\mathsf{Z}$ is independent of $\mathsf{X}$ when conditioned on
$\mathsf{Y}$.  For our purposes however, it is more convenient to eschew
the introduction of  random variables, and to consider the kernels simply
as mappings between spaces as defined above.  
Thus rather than writing a Markov chain in
terms of the random variables $\mathsf{X}$, $\mathsf{Y}$  and $\mathsf{Z}$,
$
	\mathsf{X}\stackrel{E_1}{\to} \mathsf{Y} \stackrel{E_2}{\to}
	\mathsf{Z},
$
we will write the ``chain'' as a string of experiments  operating on
spaces $X$, $Y$ and $Z$ as
$
	X\stackrel{E_1}{\expto} Y \stackrel{E_2}{\expto} Z.
$


\section{Unconstrained Information Measures --- $D$-information}
\label{sec:information-defs}
In this section we introduce the ``unconstrained'' information measure
$\I_D(E)$. The name is in contrast to the ``constrained'' family we
introduce in \S\ref{sec:Fscr-Nfrak-info}. The unconstrained information
measures subsume the classical $\phi$-divergences and their $n$-ary
generalisations (see \S\ref{sec:phi_information}).

\subsection{\texorpdfstring{$D$}{D}-information}

For a set $D\subseteq \R^n$ and an experiment $E:[n]\expto \Omega$, 
the \Def{$D$-information of $E$} is
\begin{gather}
	\Def{\I_D(E)}{ \int \sup_{d \in D}\,\rbr{ \sum_{i \in [n]} 
	d_i\cdot\diff{E_i}{\rho}}  \dv \rho},
  \label{defn:d_information}
\end{gather}
where $\rho\in\probm(\Omega)$ is a reference measure that dominates each 
of the
$(E_i)$,\footnote{It always easy to find such a $\rho$,  For example one
	may take $\rho \defeq\frac{1}{n}\sum_{i\in[n]} E_i$.} 
and $d=(d_1,\ldots,d_n)$. The definition above was first proposed by
\citet{Gushchin:2008aa} and is analogous to the
approach used by \citet{williamson2014geometry, Williamson:2022vm} where loss functions are
defined in terms of a convex set, and which forms the basis of the bridge
in \S\ref{sec:bridge}.

\begin{remark}
\label{rem:invariant-to-rho}
\normalfont
  The choice of $\rho$ is unimportant since \eqref{defn:d_information} is
  invariant to reparameterisation:
  \begin{align}
    \int \sup_{d \in D}\,\rbr{ \sum_{i \in [n]} d_i·\diff{E_i}{\rho_1}}  
       \dv \rho_1
     & =  \int \sup_{d \in D}\,\rbr{ \sum_{i \in [n]} d_i·\diff{E_i}{\rho_1}}
         \diff{\rho_1}{\rho_2}  \dv \rho_2
    \\& =  \int \sup_{d \in D}\,\rbr{ \sum_{i \in [n]} d_i·\diff{E_i}{\rho_1}
         \diff{\rho_1}{\rho_2}}   \dv \rho_2
    \\&=  \int \sup_{d \in D}\,\rbr{ \sum_{i \in [n]} d_i·\diff{E_i}{\rho_2}}
        \dv \rho_2,
  \end{align}
  for all dominating $\rho_1, \rho_2$.
\end{remark}

\begin{remark}\label{rem:d_inf_support_function}
	\normalfont
    The form of \eqref{defn:d_information} indicates we can equivalently write
  \begin{gather}
	  \Def{\I_D(E)}{ \int \sprt_{D}\left(\diff{E} \rho \right)\dv \rho},
    \label{defn:d_information_alt}
  \end{gather}
  where $\diff{E}/\rho \defeq \rbr{\diff{E_1}/\rho, \dots, \diff{E_n}/\rho}$ 
  is the vector of Radon-Nikodym derivatives, and $\sprt_{D}$ is the 
  support function of $D$ \eqref{eq:support-function-def}. 
  This suggests, using standard polar duality results,
  that the $D$-information can be viewed as an expected gauge function,
  a perspective developed in Appendix~\ref{sec:expected-gauge}.
\end{remark}

Observe that \eqref{defn:d_information} places no requirements on the
continuity of the distributions $(E_y)_{y\in [n]}$ with respect to one
another. Thus the $D$-information is more than just a multi-distribution
$\phi$-divergence; when defined as the $D$-information, Proposition
\ref{thm:fdiv_equivalence} below guarantees that $\I_{\hyp(-\phi^*)}$
agrees with $\I_\phi$ on all measures $E_1,E_2$ with $E_1\ll E_2$ and is a
natural extension to compare measures that don't have this 
absolute continuity condition\footnote{There are 
	definitions of $\phi$-divergences that hold in the general case
	\citep[p.~35]{Liese:2007ug}.  The approach we take further
	generalises to be applicable to comparisons of measures that are
	only finitely additive instead of countably additive, as explained
	by \citet{Gushchin:2008aa}, whose work was a major inspiration
for the present paper.}.  Existing generalisations of $\phi$-divergences to
$n>2$ \citep{Matusita1971, gyorfi1975f, gyorfi1978f,
Garcia-Garcia2012,Keziou:2015aa,Duchi:2017aa} are subsumed by
$D$-information.

\subsection{From \texorpdfstring{$\phi$}{phi}-divergence 
to \texorpdfstring{$D$}{D}-information}\label{sec:phi_information}

Before proceeding with a more thorough study of \eqref{defn:d_information}
we justify its introduction as a generalisation of the $\phi$-divergences.
It is convenient to slightly refine our definition of $\Phi$ as follows:
\[
    \Def{\bar{\Phi}}{\left\{\phi\st\R++\to\R,\quad\text{$\phi$ convex},  
    \quad \phi(1) = 0, 
    \quad\text{$\phi$ lsc and}\quad \R+\subseteq\tcl(\dom \phi)\right\}}.
\]
This is a very mild refinement of $\Phi$ and all $\phi$ used in the
literature on $\phi$-divergences are in fact contained in $\bar\Phi$.
Observe that demanding $\phi$ be a  proper function to $\R$ defined on all
of $\R_{>0}$ implies that $\R+\subseteq\tcl(\dom \phi)$.  Assuming lower
semi-continuity is a mere convenience since one can enforce it by taking
closures, and, as we shall see, the information functionals will not change
in this case since they are expressible in terms of support functions of
the epigraph of functions related to $\phi$, which remain invariant under
taking closures of the sets concerned. In any case, $\tcl f$ and $f$
coincide on $\relint\dom f$ 
\citep[Proposition B.1.2.6]{hiriarturruty2001fca}.  
If we simply require that $f(x)<\infty$ for all
$x\in(0,\infty)$ then lower semicontinuity and the claim re domain follow
as logical consequences.

Suppose $\mu,\nu\in\probm(\Omega)$, with a common dominating measure $\rho$. 
Choose some $\phi\in\bar\Phi$.
Then the $\phi$-divergence \eqref{def:fdiv_fmi} has the following
representation using the perspective function $\breve\phi$,\footnote{This
	observation is due to \citet{Gushchin:2008aa}.}
\begin{align}
  \I_\phi(\mu,\nu)
   & = \int \phi\rbr{\diff{\mu}{\nu}}\dv \nu
   \label{eq:standard-phi-formula}
  \\& = \int \breve\phi\rbr{\diff{\mu}{\nu}, 1} \diff \nu\rho\dv \rho
  \\& = \int\breve\phi\rbr{\diff{\mu}{\nu}\diff \nu\rho,\diff \nu\rho}\dv\rho
  \\& = \int \breve\phi\rbr{\diff{\mu}{\rho}, \diff \nu\rho}\dv \rho.
       \label{eq:perspective_f_divergence}
\end{align}
Equation \eqref{eq:perspective_f_divergence} 
is symmetric in $\mu$ and $\nu$, in contrast to
\eqref{eq:standard-phi-formula}, with any intrinsic asymmetry relegated to
the choice of sublinear function $\breve\phi$.  By the same argument used
in Remark \ref{rem:invariant-to-rho}, the choice of $\rho$ does not matter.
Observe that upon substituting the definition of the perspective into
(\ref{eq:perspective_f_divergence}) we obtain the formula
\[
    \I_\phi(\mu,\nu)=\int \diff \nu\rho\cdot  
    \phi\rbr{\frac{\dv\mu/\dv\rho}{\dv\nu/\dv\rho}}\dv\rho,
\]
as recently observed in \citep[Remark 19]{Agrawal2021}, and which of course
remains invariant to the the choice of $\rho$.

\begin{remark}
	\normalfont
  It is a common result in nonsmooth analysis  (due to Hörmander
  \citep[Corollary 1.81, p.~56]{Penot2012Calculus}) that the
  mapping taking a set to its support function, $D\mapsto \sprt_D$, is an
  injection from the family of closed convex subsets to the set of
  positively homogeneous functions that are null at zero. Thus it is
  natural, as well as meaningful for our subsequent analysis, to
  parameterise \eqref{eq:perspective_f_divergence} by a convex set as in
  \eqref{defn:d_information} or \eqref{defn:d_information_alt}.  That is,
  given $\phi$, we will work with the convex set $D\in\cvx(\R^2)$ such that
  $\breve{\phi}=\sprt_D$; an explicit formula for such a $D$ in terms of
  $\phi$ is provided in Proposition \ref{thm:fdiv_equivalence} below.
\end{remark}

Since we will be considering $n$-ary extensions of $\I_\phi$, it is
convenient to number the measure arguments and stack them into a vector $E
\defeq (E_1,E_2)$, in which case the pair $(E_1,E_2)$ may interpreted,
equivalently, as a binary experiment $[2]\expto \Omega$.

\begin{proposition}\label{thm:fdiv_equivalence}
  Suppose $E:[2]\expto \Omega$ satisfies $E_1 \ll E_2$ and $\phi\in\bar\Phi$.
  Let 
  \begin{gather}
	  \label{eq:D-phi-def}
	  \Def{D_\phi}{\hyp(-\phi^*) \subseteq\R^2}.
  \end{gather}
  Then
  \begin{gather}
    \I_\phi(E_1,E_2) = \I_{D_\phi}((E_1,E_2)).
  \end{gather}
\end{proposition}
\begin{proof}
  The assumptions on $\phi$ ensure that it is closed. We have
  \begin{align}
    \I_\phi(E_1,E_2)
     & = \int \phi\rbr{\diff {E_1}{E_2}} \dv E_2\\
    &= \int \cl\breve\phi\rbr{\diff {E_1}{\rho}, \diff
        {E_2}{\rho}}\dv\rho \label{eq:perspective_f_divergence_2}\\
    &=\int\sprt_{\hyp(-\phi^*)}\rbr{\diff {E_1}{\rho},
      \diff {E_2}{\rho}}\dv\rho \label{eq:persp_hypograph_relation}\\
    &= \I_{\hyp(-\phi^*)}(E),
  \end{align}
    where \eqref{eq:perspective_f_divergence_2} holds because  a closed
    convex function $\phi$ and the closure of its perspective satisfies
    $\phi = \cl\breve\phi(\marg, 1)$, \eqref{eq:persp_hypograph_relation}
    follows from the the relationship $\cl\breve{\phi}(x,t)= \sprt_{\epi
    \phi^*}(x,-t)$ \citep[Proposition 1.2.1, p.~214]{hiriarturruty2001fca}, 
    and it is easy to verify that
    $\sprt_{\epi \phi^*}(x,-t) = \sprt_{\hyp(-\phi^*)}(x,t)$. 
\end{proof}

\begin{proposition}
\label{prop:phi-star-in-D}
Suppose $\phi\in\bar\Phi$ then
	$\hyp(-\phi^*)\in\cvxrec\rbr{\R^n,\R_{\leq0}^n}$ and
	$\sigma_D(1_2)=0$. 
\end{proposition}
\begin{proof}
	The Fenchel conjugate is always closed and convex, thus
	$\hyp(-f^*)$ is closed and convex. Let $A$ be the linear operator
	that flips the sign of the last element a vector $x\in\R^n$;
	$
	\Def{Ax}{(x_1,\dots,x_{n-1},-x_n)}.
	$
	Then \citep[Proposition 2.1.11, p.~31]{Auslender2003}   implies
	that
	\[
		\hyp(-\phi^*) = A\epi(\phi^*)\mbox{\ \  and\ \ }
		\rec (\hyp(-\phi^*)) = A\rec(\epi \phi^*).
	\]
	\citet[Theorem 2.5.4, p.~55]{Auslender2003} show  $\rec(\epi f^*) =
	\epi(\sprt_{\dom f})$.  Since $\sprt_{\dom \phi}$ is 1-homogeneous
	and closed, its epigraph is a closed cone, thus it is equal to its
	recession cone \citep[Proposition 2.1.1, p.~26]{Auslender2003},
	that is, $\rec(\epi \sprt_{\dom \phi}) = \epi(\sprt_{\dom\phi})$.
	By assumption $\cl(\dom\phi)\supseteq\R_{\geq 0}$, thus
	$\sprt_{\dom\phi} = \sprt_{\cl(\dom\phi)}
	=\ind_{-(\cl(\dom\phi))^+}$, where, recall, the ${}^+$ denotes the
	dual cone \eqref{eq:dual-cone-def}.  Since $\phi\colon\R_{\ge
	0}\rightarrow\R$ is presumed to be defined (finite)  on the whole
	of $\R_{\ge 0}$, $\cl(\dom\phi)\supseteq \R_{\geq0}$ and thus  we
	have $\cl(\dom\phi)^*\subseteq \R_{\geq0}$; thus
	$\epi(\ind_{\R^n_{\leq 0}}) \subseteq \R_{\leq 0} \times \R_{\geq
	0}$. This gives us
	\begin{gather}
		\rec\rbr{\hyp(-\phi^*)} = A\rec(\epi(\sprt_{\dom \phi})) = 
		A\epi(\sprt_{\dom\phi}) = A(\R_{\leq 0} \times \R_{\geq 0}) = 
		\R_{\leq 0}^2,
	\end{gather}
	which shows $\hyp(-\phi^*)\in\cvxrec\rbr{\R^n,\R_{\leq0}^n}$.
	Finally we have that
	$\sigma_D(1_2)=\breve\phi(1,1)=\phi(1)=0$ by assumption on $\phi$.
\end{proof}
We also have the following converse result:
\begin{proposition}\label{prop:phi-D-in-Phi}
Let   $D\in\cvxrec(\R^2,\R_\le^2)$ with $\sigma_D(1_n)=0$. Let
	$\Def{\phi_D(x)}{\sigma_D((x,1))}$.
	Then $\phi_D\in\bar{\Phi}$.
\end{proposition}
\begin{proof}
	Support functions are convex and thus it is immediate that $\phi_D$
	is too. We have $\phi_D(1)=\sigma_D((1,1))=0$ by assumption.
	Since $\rec D=\R_{\le 0}^2$  we have $\dom\sigma_D=\R_{\ge 0}^2$
	and thus $\dom\phi_D=[0,\infty)$.
\end{proof}

Since Proposition \ref{thm:fdiv_equivalence} shows every $\phi$-divergence 
corresponds to a $D$-information, it is natural then to ask 
which $D$-informations correspond to $\phi$-divergences.  
Similarly to Proposition \ref{thm:fdiv_equivalence} we may obtain, 
from any $D\subseteq \R^2$ a convex, lower semicontinuous 
function $\phi_D\colon \R\to\Rx$ by the mapping $D\mapsto \sprt_D(\marg, 1)$.
Ensuring that this function is finite on $\R+$ and normalised 
appropriately to be consistent with \eqref{eq:phi_desiderata} is more subtle.

\begin{proposition}\label{prop:set_normalisation}
  Suppose $D\subseteq\R^n$ is nonempty and closed convex. 
  Then we have $\sprt_D ≥ 0$ if and only if $0\in D$. 
  Assume $0\in D$, and let $ Z\defeq\setcond{c1_n\in\R^n}{c>0}$. 
  Then  $\sprt_D$ is minimised with minimum value $0$ along 
  the $Z$ ray if and only if $D\subseteq \halfspace^{≤}_{1_n}$.
\end{proposition}
\begin{proof}
  The common result \citep[theorem C.3.3.1]{hiriarturruty2001fca} that 
  $A\subseteq B$ if and only 
  if $\sprt_A≤\sprt_B$ with $A = \set{0}$  easily shows the first claim.
  For the remainder of the proof assume $0\in D$.  Whence
  \begin{align}
    D\subseteq \halfspace^{≤}_{1_n} \iff \forall{d\in D} 
    \inp{d,1_n} ≤ 0 \iff \sup_{d\in D} \inp{d,1_n} ≤ 0 \iff \sprt_D(1_n) ≤ 0.
  \end{align}
  Since $\sprt_D ≥ 0$ (owing to the assumption $0\in D$), 
  we must have $\inf\sprt_D = \sprt_D(1_n) = 0 $. 
  Positive homogeneity of $\sprt_D$ implies this holds along the ray $Z$ too.
\end{proof}

\begin{corollary}
  Suppose $D\subseteq\R^n$ is nonempty and closed convex. 
  Then  $\sprt_D$ is minimised along the $Z$ ray if and 
  only if $D - \subd \sprt_D(1_n)\subseteq \halfspace^{≤}_{1_n}$.
\end{corollary}

Observe that \eqref{defn:d_information} places no requirements on the
continuity of the distributions $(E_y)_{y\in Y}$ with respect to one
another. Thus the $D$-information is more than just a multi-distribution
$\phi$-divergence; when defined as the $D$-information, Theorem
\ref{thm:fdiv_equivalence} guarantees that $\I_{\hyp(-\phi^*)}((E_1,E_2))$
agrees with $\I_\phi(E_1,E_2)$ on all measures $E_1,E_2$ with $E_1\ll E_2$,
and it is a natural extension to compare measures that don't have this
absolute continuity condition\footnote{There are more
	general definitions of $\phi$-divergences that hold in the general
	case; see e.g.~\citep[p.~35]{Liese:2007ug}.  The approach we
	take further generalises to be applicable to comparisons of
	measures that are only finitely additive instead of countably
	additive, as explained by \citet{Gushchin:2008aa}, whose work
was a major inspiration for the present paper.}.  Most existing
generalisations of $\phi$-divergences to $n>2$ \citep{Matusita1971,
	gyorfi1975f, gyorfi1978f, Garcia-Garcia2012, Keziou:2015aa} are
	subsumed by $D$-information; the one exception
	\citep{Birrell:2022} is discussed in Appendix
	\ref{sec:precursors}.

Thus the $D$-informations that correspond to a normalised
$\phi$-divergence (with $\phi$ strictly convex) 
are those strictly convex $D\in\cvxrec({\R^n,
\R_{\leq0}^n})$ which lie in the half space with outer normal
vector $1_n$ and pass through the origin at their boundary.  This
normalisation corresponds to the well known fact that
$\phi$-divergences are insensitive to affine offsets:   
\begin{proposition}
\label{prop:affine-offset}
	Suppose $\phi\in\bar\Phi$ and $c\in\R$. Let
	$\Def{\phi_c(x)}{\phi(x)+c(x-1)}$.  Then
	\begin{align}
		\I_{\phi_c} &=\I_\phi\label{eq:phi-c-0}\\
		\phi_c^*(x)&=\phi^*(x-c)+c\label{eq:phi-c-1}\\
		\breve{\phi}_c(s,t)&=\breve{\phi}(s,t)+c(s-t)
		    \label{eq:phi-c-2}\\
		\sigma_{D_{\phi_c}}(s,t)
			&=\sigma_{D_\phi}(s,t)+c(s-t)
			  \label{eq:phi-c-3}\\
		D_c &= D+ \{(c, -c)'\}. \label{eq:phi-c-4}
	\end{align}
\end{proposition}
Observe that transforming $D$ to $D_c$ corresponds to ``sliding'' $D$ along
the supporting hyperplane $\left\{x\in\R^2 \st \langle x,1_2\rangle
=0\right\}$, i.e.~ the boundary of $\halfspace_{1_2}^\le$.
\begin{proof}
	Substituting $\phi_c$ into \eqref{def:fdiv_fmi} gives
	\eqref{eq:phi-c-0}.  Equation \eqref{eq:phi-c-1} follows from
	\citep[Proposition E.1.3.1 (i) and (vi)]{hiriarturruty2001fca}.
	Equation \eqref{eq:phi-c-2} follows by substitution into the
	definition of the perspective. Equation \eqref{eq:phi-c-3} follows
	from \eqref{eq:phi-c-2} by the fact that the perspective of $\phi$
	is the support function of $D_\phi$, and \eqref{eq:phi-c-4} follows
	from the additivity of support functions under Minkowski sums
	\citep{Schneider1993} and the support function of a singleton being
	a linear function \citep{hiriarturruty2001fca}.
\end{proof}
\begin{lemma}
\label{prop:sliding-D}
  Suppose $D\subseteq\R^n$ and let $E:[n]\expto\Omega$ be an experiment. 
  Suppose $p\in\R^n$ is such that $\langle p,1_n\rangle =0$. 
  Let $\Def{D_p}{D +\{p\}}$. Then $\I_{D_p}(E)=\I_D(E)$.
  \end{lemma}
  \begin{proof}
      From \eqref{defn:d_information_alt} we have
      \begin{align}
          \I_{D_p}(E) &=\int\sigma_{D_p}\left(\diff{E}{\rho}\right)\dv\rho\\
          &=\int\left(\sigma_{D}\left(\diff{E}{\rho}\right) 
              +\left\langle p,\diff{E}{\rho}\right\rangle\right)\dv\rho\\
          &=\I_D(E)+\int\left\langle p,\diff{E}{\rho}\right\rangle\dv\rho\\
          &=\I_D(E)+\left\langle p,\int\diff{E}{\rho}\dv\rho\right\rangle\\
          &=\I_D(E)+\langle p,1_n\rangle\\
          &=\I_D(E),
      \end{align}
      where the second equality follows from additivity of support
      functions of Minkowski sums, and the fact that
      $\sigma_{\{p\}}(x)=\langle p,x\rangle$.
  \end{proof}
  A special case of this result is when $n=2$ and $p=(c,-c)$ which
  corresponds to the situation of Proposition \ref{prop:affine-offset},
  showing that translating $D$ in the manner of Lemma
  \ref{prop:sliding-D} corresponds to the classical result that an affine
  offset to $\phi$ does not change $\I_\phi$.

\begin{remark}\label{rem:zero-in-boundary}
	\normalfont
With this result it is clear that we can always canonically assume that for
any $D$ such that $\sigma_D(1_n)=0$, we have $0_n\in\tbd D$. To see this,
suppose $0_n\not\in\tbd D$, and denote by
$\Def{s}{\operatorname{D}\sigma_D(1_n)}$, the support point of $D$ in
direction $1_n$. Then using  $v=-s$ in the above proposition to determine
$D_v$ ensures $\I_{D_v}=\I_D$ and that $0_n\in\tbd D_v$.  Requiring
$0_n\in\tbd D$ and $\sigma_D(1_n)=0$ corresponds, in the case that $n=2$,
to choosing the affine offset for $\phi$ such that $\phi$ is everywhere
non-negative.
\end{remark}

\begin{proposition}\label{prop:tni-D}
  Let $D\in\cvxrec(\R^n,\R_{\leq0}^n)$.
  Then $\I_D(E^{\mathrm{tni}})=0$ for all totally
  non-informative experiments $E^{\mathrm{tni}}$ if and only if
  $ D \subseteq\halfspace_{1_n}^\le= \setcond{x\in\R^n}{ \inp{ x , 1_n}  \leq 0}.$
\end{proposition}
\begin{proof}
  Let $D$ be such that for all  totally non-informative experiment
  $E^{\mathrm{tni}}$, $\I_D(E^{\mathrm{tni}})=0.$ This means that 
  \begin{gather}
  \E_{\mu}\left[\sprt_D(c(x)1_n) \right] = \E_{\mu}\left[c(x) \sprt_D(1_n)
  \right] = \sprt_D(1_n) \int_X c(x) d\mu(x)= 0
  \end{gather}
  for all measures $\mu,$
  and all functions $c \colon \R \to \R_+.$ Hence, $\sprt_D(1_n)=0.$ 
  By definition of the support function,   $\sprt_D(1_n)=0$ means that the
  hyperplane $\{x\st\langle x,1_n\rangle=0\}$ supports $D$ and thus
  $D\subseteq\{x\st\langle x,1_n\rangle\le 0\}$.
  Conversely, if  $ D \subseteq \{x\in\R^n\st \inp{ x , 1_n}  \leq 0\},$
  $\forall d \in D, \inp{ d, 1_n}  \leq 0,$ and so $\sprt_D(1_n)=0,$ which
  gives $\I_D(E^{\mathrm{tni}})=0$ for all totally non-informative
  experiments $E^{\mathrm{tni}}$.
\end{proof}
In light of the above arguments, we define the class of such normalised $D$ 
by\footnote{These sets are also called ``comprehensive'' 
	(``downward'' and convex); see
	\citep{martinez2002downward}.}
\[
	\Def{\Dfrak^n}{\left\{D\in\cvxrec(\R^n,\R_{\leq0}^n)
	\st  \sigma_D(1_n)=0\right\}}.
\]
Observe that $\sigma_D(1_n)=0$ and  $\rec
D=\R_{\le 0}^n$ together imply that $D\subseteq \lev_{\leq 0}\inp{\marg, 1_n}$.
Given Remark \ref{rem:zero-in-boundary}, 
we could always restrict ourselves to 
\[
	\Def{\Dfrak_0^n}{\left\{D\in\Dfrak^n \st  0_n\in\tbd D\right\}}.
\]
Although many of the results below hold for more general choices of $D$,
one loses nothing (in terms of the expressive power of $\I_D$) in
restricting $D$ to $\Dfrak^n$ or indeed $\Dfrak_0^n$.
In the case where $n=2$, so $Y=\cbr{1,2}$, for $D\in\Dfrak^n$, the
corresponding function $\phi$ such that $\I_\phi=\I_D$ can be obtained as
the mapping $\phi_D\colon x\mapsto \sprt_D((x, 1))$  and one is guaranteed
that $\phi_D\in\bar{\Phi}$ (Proposition \ref{prop:phi-D-in-Phi}).
Some further observations on the relationship between $D$-information and
$\phi$-information are given in Remark \ref{rem:witness-phi}.

\subsection{Properties of \texorpdfstring{$D$}{D}-information} 
\label{sec:properties-of-D-information}

Since $D_1\subseteq D_2\ \Longleftrightarrow\ \sigma_{D_1}\le \sigma_{D_2}$,
\eqref{defn:d_information_alt} immediately gives that 
$
	D_1\subseteq D_2 \ \Longleftrightarrow\  \I_{D_1}\le \I_{D_2}.
$
The $D$-information is insensitive to
certain operations on $D$:  
taking closed convex hulls;  and taking
Minkowski sums with the negative orthant:
 \begin{lemma}\label{lem:dinf_conv_closure}
       Suppose $D\subseteq\R^n$ is closed and let $E:[n]\expto\Omega$ be an
       experiment.  Then
       \begin{gather}
         \I_{D}(E) = \I_{\clco(D)}(E) = \I_{\clco(D) + \R-^n}(E).
       \end{gather}
 \end{lemma}
 \begin{proof}
       Using some elementary properties of the support function
       \citep{hiriarturruty2001fca,Auslender2003} $\sprt_D =
       \sprt_{\clco D}$. Appealing to Definition
       \ref{defn:d_information_alt} this shows the first equality.  In
       order to
       prove the second we use the fact that $\sprt_C = \ind_{C^\dc}$,
       where $C$ is a cone and $C^\dc$ is its dual cone
       \eqref{eq:dual-cone-def}.
       Thus
       \[
         \sprt_{\clco(D) + \R-^n} =
         \sprt_{\clco(D)}+\sprt_{\R+^n} =
         \sprt_{\clco D}  + \ind_{\R+^n},
       \]
       where the last step is a consequence of $(\R+^n)^\dc=\R+^n$
       \citep[p.~49]{hiriarturruty2001fca}. Since the function $\diff
       E/\rho$ maps into $\R^n_{\geq 0}$, appeal to the alternate definition
       \eqref{defn:d_information_alt} completes the proof.
 \end{proof}

 \begin{proposition}\label{thm:quotient_space_isomorphism}
       The $D$-information induces a quotient space on the closed convex
       sets $A,B\subseteq\R^n$ where $\rec A \subseteq \R-^n$ and $\rec B
       \subseteq \R-^n$ via the equivalence relation
       \begin{gather}
         A\sim_{\I} B \iff \I_{A} = \I_{B}.
       \end{gather}
       This quotient space is isomorphic to $\cvxrec({\R^n, \R-^n})$.
 \end{proposition}
 \begin{proof}
       By hypothesis $D\subseteq \R^n$ and $\rec D \subseteq \R-^n$.  Thus
       $\rec\rbr{D + \R-} = \rec D + \R- = \R-$; the inclusion follows from
       \citep[Theorem 2.3.4, p.~39]{Auslender2003}.  Finally
       we note $\clco(D) \in \cvx(\R^n)$, which, together with Lemma
       \ref{lem:dinf_conv_closure} completes the proof.
 \end{proof} 

\begin{remark}
\normalfont
 Proposition \ref{thm:quotient_space_isomorphism} has a simple
 interpretation since for all bounded subsets $D\subseteq\R^n$, $\rec D =
 \cbr{0}$; and thus the equivalence relation applies to these in addition to
 any set (unbounded) that recesses in directions $R\subseteq\R^n_{\leq0}$.
 Thus $\cvxrec(\R^n,\R-^n)$ is the natural parameter space for $\I_D$.
 \end{remark}

Some $\phi$ divergences (e.g.~variational) are always bounded, and others
(e.g.~Kullback-Leibler) are not. There is a simple characterisation of
when $\I_D$ is guaranteed to be bounded:
\begin{proposition}\label{prop:limID}
	Suppose $D\in\cvxrec({\R^n, \R-^n})$. Then
	\[
		\sup_{E\colon [n]\expto X} \I_D(E)<\infty 
	\]
	if and only if there exists some
	$\alpha\in\R^n$ such that $D\subset \R_{\le 0}^n +\{\alpha\}$.
\end{proposition}
\begin{proof}
	We first show that $\sup_E \I_D(E)<\infty$   if and only if
	$\sigma_D(x)<\infty$ for all $x\in\R_{\ge 0}^n$.  Recall
	$\I_D(E)=\int \sprt_{D}\left(\varfrac{\dv {E}}{ \dv\rho} \right)\dv \rho$.  If
	there exists $x^*\in\R_{\ge 0}^n$ such that $\sigma_D(x^*)=\infty$,
	then we can always choose $E^*$ such that $\left(\dv{E^*}/\dv{\rho}
	\right)(z) = c x^*$ for some $c>0$ for all $z$, and thus
	$\I_D(E^*)=\infty$. Furthermore, if $\I_D(E^*)=\infty$ for some
	$E^*$ then it must be the case that for at least one $z$, we have
	$\sigma_D((\dv{E^*}/ \dv\rho)(z))=\infty$.  Conversely if
	$\sigma_D(x)<\infty$ for all $x$ then there is no way $\I_D(E)$ can
	be made infinite by choice of $E$.   Furthermore, if
	$\I_D(E)<\infty$ for all $E$ then $\I_D(E^*)$ for $E^*$ such that
	$(\varfrac{\dv{E^*}}{\dv \rho})(z)=c x $ for arbitrary $x\in\R_{\ge 0}^n$ and
	some constant $c$ (recall $\sigma_D$ is 1-homogeneous).  It thus
	follows that $\sigma_D(x)$ must not be infinite for all $x$.

	We now show $\sigma_D(x)<\infty$ for all $x\in\R_{\ge 0}^n$ if and
	only if $D\subset \R_{\le 0}^n +\{\alpha\}$.  Suppose $D\subset
	\R_{\le 0}^n +\{\alpha\}$. Then for all $x\in\R_{\ge 0}^n$,
	$\sigma_D(x)\le \sigma_{\R_{\le 0}^n}(x)+\sigma_{\{\alpha\}}(x) =
	0+\langle \alpha,x\rangle<\infty$.   Conversely, if
	$\sigma_D(x)<\infty$ for all $x\in\R_{\ge 0}^n$ then
	$\sigma_D(\e_i)<\infty$ for $i\in[n]$ where $\e_i$ is the $i$th
	canonical basis vector.  But $\sigma_D(\e_i)<\infty \Rightarrow
	\sup_{y\in D}\langle y,\e_i\rangle<\infty \Rightarrow D\subset
	\halfspace_{i,\alpha_i}$ for some $\alpha_i\in\R$, where
	$\halfspace_{i,\alpha_i}=\{x\st\langle x,\e_i\rangle \le
	\alpha_i\}$ is the halfspace with normal $\e_i$ and offset
	$\alpha_i$. Since this holds for all $i\in[n]$ we have that
	$D\subset \bigcap_{i\in[n]} \halfspace_{i,\alpha_i}=\R_{\ge
	0}^n+\{\alpha\}$, with $\alpha=(\alpha_1,\ldots,\alpha_n)$.
\end{proof}

\begin{remark}
	\normalfont
	\label{rem:bss}
	We can express the Blackwell-Sherman-Stein theorem  \citep[section
	3.2.2]{Ginebra:2007} in terms of $\I_D$.   Say one experiment
	$E\colon [n]\expto X$ is \Def{better than} $F\colon [n]\expto X$,
	and write $\Def{E \succcurlyeq F}$, if there exists a Markov kernel
	$T\colon X\expto X$
	such that $F=ET$; that is, experiment $F$ can be obtained from
	experiment $E$ by applying some corruption kernel $T$. 
	The theorem states: 
	\begin{gather}
		\label{eq:bss-convex-criteria}
	E \succcurlyeq F \ \Longleftrightarrow\ 
		\int f\left(\diff E \rho\right)\dv \rho \ \ge\ 
		\int f\left(\diff F \rho\right) \dv\rho,\ \ \ 
		\forall f\colon\R_{\ge 0}^n\rightarrow\R, \ f \mbox{\ convex}.    
	\end{gather}
	(As
	usual, the choice of dominating measure $\rho$ does not matter).
	We now argue that we can replace $f$ by $\sigma_D$ with
	$D\in\Dfrak^n$. Since $\diff E \rho (x), \diff F \rho (x)\in\R_{\ge
	0}^n$ for all $x$, it suffices to ensure $\dom \sigma_D = \R_{\ge
	0}^n$ which is guaranteed by the fact that $\rec D=\R_{\le 0}^n$.
	Since $f$ appears on both sides of \eqref{eq:bss-convex-criteria},
	an additive offset is cancelled, and thus we can always subtract
	$\sigma_D(1_n)$ from both sides which is tantamount to assuming
	$\sigma_D(1_n)=0$. Thus we can replace 
	\eqref{eq:bss-convex-criteria} by
	\begin{gather}
	E \succcurlyeq F \ \Longleftrightarrow\ 
		\I_D (E)\ \ge\ \I_D(F), \ \ \ \forall D\in\Dfrak^n.
	\end{gather}
	That is, $E$ is better than $F$ if and only if, \emph{for all} $D$,
	the $D$-information of $E$ is greater than or equal to the
	$D$-information of $F$; one cannot compare $E$ and $F$ in the
	absolute sense of $\succcurlyeq$ by using only \emph{one} measure of
	information.
\end{remark}

\section{The Bridge between Information and Risk}
\label{sec:bridge}
Having introduced the $D$-information, in this section we show its
connection to the Bayes risk, and present the corresponding information processing
equality.  

\subsection{\texorpdfstring{$D$}{D}-information and Bayes Risk}
\label{subsec:bridge}

Classically, a \emph{loss function} is a mapping
$\ell:\probm([n])\times[n]\to\bar\R_+$, where the quantity $\ell(\mu, y)$
is to be interpreted as the penalty incurred when predicting
$\mu\in\probm([n])$ under the occurrence of the event $y\in[n]$. A loss
function is said to be \Def{proper} if the expected loss is minimised by
predicting correctly, and \Def{strictly proper} if it is minimised by
predicting precisely\footnote{See \citep{McCarthy1956,Buja:2005,
Reid:2010,williamson2014geometry, Williamson:2022vm} for further background
and history of proper losses.}. That is, for all $\mu\in\probm([n])$
\begin{gather}
  \mu \in\arginf_{\nu\in\probm([n])} \E_{\Ysf\sim\mu}\sbr{\ell(\nu, \Ysf) },
  \quad\text{and}\quad
  \set{\mu} =\arginf_{\nu\in\probm([n])} \E_{\Ysf\sim\mu}\sbr{\ell(\nu,
	  \Ysf) },
\end{gather}
respectively. Considering a product space $\Omega\times[n]$ and measures
$\mu\in\probm(\Omega\times[n])$, $\nu\in\probm([n])$, we introduce two
classical quantities, the \Def{Bayes risk}, and \Def{conditional Bayes
risk}:
\begin{align}
	& \Def{\brisk_\ell(\mu)}{ \inf_{f\in\measf(\Omega, \probm([n]))}
	\E_{(\Xsf,\Ysf)\sim\mu}\sbr{ \ell(f(\Xsf), \Ysf)  } }\label{defn:bayes_risk}
  \intertext{and}
  & \Def{\cbrisk_\ell(\nu)}{ \inf_{\nu'\in\probm([n])} \E_{\Ysf\sim\nu}\sbr{
  \ell(\nu', \Ysf)  }}.
\end{align}
These are related by
\begin{gather}
  \brisk_\ell(\mu) = \E_{\Xsf\sim \mu_\Xsf}\sbr{ \cbrisk_\ell(\mu_{\Ysf|\Xsf})},
  \label{eq:bayes_cbayes_expectation}
\end{gather}
where $\mu_\Xsf$ is the law of $\Xsf$ and $\mu_{\Ysf|\Xsf}$ is the
conditional distribution of $\Ysf$ given $\Xsf$.

We stack $\ell$ into a vector over its second argument:
$\Def{\ell(\mu)}{\rbr{ \ell(\mu,1), \dots, \ell(\mu,n) }}.$ The
\Def{superprediction set}\footnote{See
\citep{KalnishkanVovkVyugin2004,Dawid2007,
crankoAnalyticApproachStructure2021,Williamson:2022vm} 
for uses of the superprediction set.}
associated to
$\ell$ is
\begin{gather}
	\Def{\super(\ell)}{\setcond*{x\in\R^n }{
    \exists{\mu\in\probm([n])}  x-\ell(\mu)  \in \R+^n
  }}.
  \label{def:superprediction_set}
\end{gather}
The superprediction set is the set of all ``superpredictions'' --- points
``north-east'' of the image of the loss $\ell(\probm([n]))$:
\begin{gather}
  \super(\ell) = \ell(\probm([n])) + \R+^n.\label{eq:super_sum}
\end{gather}
There are relationships between properties of $\ell$ and the geometry 
of $\super(\ell)$. For example:
\begin{enumerate}
  \item $\ell$ is proper  only if $\super(\ell)$ is convex when 
      $\ell$ is continuous 
      \citep[Theorem 4.13]{crankoAnalyticApproachStructure2021}.
  \item $\ell$ is $\eta$-mixable \citep{Vovk:1995} if $e_\eta(\super(\ell))$
	  is convex, where for $\eta>0$, $e_\eta\colon\R^n\ni x\mapsto
	  (e^{-\eta x_1},\ldots,e^{-\eta x_n})$ and for $S\subset\R^n$,
	  $e_\eta(S)=\{e_\eta(s)\st s\in S\}$. Equivalently, a loss $\ell$
	  is mixable if and only if $\super(\ell)$ slides freely in
	  $\super(\ell_{\mathrm{log}})$, the superprediction set for
	  log-loss \citep{Cabrera:2023aa}.
\end{enumerate}

\begin{remark}\label{rem:bayes_risk_support_function} \normalfont
	The conditional Bayes risk of w.r.t.~a loss $\ell$  has the following
	representation using the support function and superprediction set:
  \begin{align}
    \cbrisk_\ell(\mu)
     & =\inf_{\nu'\in\probm([n])} \E_{\Ysf\sim\nu}\sbr{ \ell(\nu', \Ysf)  }
    \\&= \inf_{\nu'\in\probm([n])} \inp{\ell(\nu'), \nu}
    \\&= \inf_{f\in\ell(\probm([n])) + \R+^n} \inp{f, \nu}
    \\&= -\sup_{f\in\ell(\probm([n])) + \R+^n} \inp{-f, \nu}
    \\&= -\sup_{f\in -(\ell(\probm([n])) + \R+^n)} \inp{f, \nu}
    \\&= -\sprt_{-\super(\ell)}(\mu).
  \end{align}
\end{remark}

Since the Bayes risk and $D$-information can both be written in terms of a
support function,  it is unsurprising that there is a relationship between
them, and in fact it is simple.  In order to demonstrate this, we need some
technical results first.

\begin{lemma}[\protect{\citep[Theorem 14.60]{RockafellarWets2004}}]
  \label{lem:infimum_interchange}
  Suppose $(\Omega, \Sigma)$ is a measurable space and let $\cal
  F\subseteq\measf(\Omega,\R^n)$ be \emph{decomposable} relative to a
  sigma-finite measure $\rho$ on $\Sigma$.  Let $\psi:
  \Omega\times\R^n\to\Rx$ be a \emph{normal integrand}\footnote{The
	  technical
	  terms ``normal integrand'' and ``decomposable'' are defined by
	  \citet[Definition 14.27, 14.59]{RockafellarWets2004}, to 
	  which we refer the reader for
	  details. },
  and let $\Def{I_\psi(f)}{\int \psi(x,f(x))\rho(\dv x)}$.
  If $I_\psi\not\equiv \infty$ on $\Omega$ then
  \begin{gather}
    \inf_{f\in\cal F} I_\psi(f) =
    \int \rbr{\inf_{f\in\R^n} \psi(x,f)
    } \rho(\dv x).\label{eq:rfw}
  \end{gather}
\end{lemma}
\begin{lemma}\label{lem:integral_sprt_function_trick}
  Let $(\Omega,\Sigma)$ be a measurable space, and $\rho$, a
  sigma-finite measure on $\Sigma$. Let $D\subseteq\R^n$ be
  nonempty, closed and measurable. Let $k: \Omega\times \R^n\to\Rx$ be such that
  $k(\marg, d)$ is measurable for all $d\in\R^n$ and  $-k(x,\marg)$ is
  convex and lower semi-continuous for all $x\in \Omega$.  Then
  \begin{gather}
    \sup_{f\in\measf(\Omega,D)}
    \int k(x, f(x)) \dv\rho(x) = \int \sup_{d\in D}k(x, d) \dv \rho(x).
  \end{gather}
\end{lemma}

\begin{proof}
  In order to apply Lemma \ref{lem:infimum_interchange}, let
  $\Def{\psi(x,d)}{\ind_{D}(x) - k(x,d)}$. Since $\psi$ is the sum of the
  indicator function  of a closed measurable set and an appropriately
  measurable, lower semicontinuous map, it is normal \citep[Proposition
  14.39]{RockafellarWets2004}. The
  collection $\measf(\Omega,\R^n)$ is trivially decomposable. Therefore
  \begin{align}
    \sup_{f\in\measf(\Omega, D)}\int k(x,f(x)) \rho(\dv x)
     & \overset{\hphantom{\eqref{eq:rfw}}}{=}
    \sup_{f\in\measf(\Omega,\R^n)}
    \int\rbr{k(x,f(x)) -(\ind_{D}\circ f)(x)}\rho(\dv x)
    \\&\overseteqref{eq:rfw}{=} \int \sup_{d\in \R^n}\rbr{k(x,d) -
      \ind_{D}(d)} \rho(\dv x)
    \\&= \int \sup_{d\in D}k(x,d) \rho(\dv x).\qedhere
  \end{align}
\end{proof}

\begin{lemma}\label{lem:brisk_with_superprediction_set}
  Suppose $\Omega$ is a standard Borel space, 
  $\ell\in\measf(\Omega\times[n],\R)$, then
  \begin{gather}
    \forall{\mu\in\probm(\Omega\times[n])}
    \brisk_\ell(\mu)
    = \inf_{f\in\measf(\Omega,\super(\ell))}\int f \dv \mu.
  \end{gather}
\end{lemma}
\begin{proof}
  Since $\Omega$ is a standard Borel space, so is $\Omega\times[n]$ and we
  have $\mu = \mu_\Xsf\times \mu_{\Ysf|\Xsf}$.  Let
  $\Def{S}{\ell(\probm([n]))}$.
  Then
  \begin{align}
    \brisk_\ell(\mu)
     & =\inf_{f\in\measf(\Omega, \probm([n]))} 
          \E_{(\Xsf,\Ysf)\sim\mu}\sbr{ \ell(f(\Xsf), \Ysf)  }
    \\& =\inf_{f\in\measf(\Omega, \probm([n]))} \int
	    \inp{\ell(f(x)), \mu_{\Ysf|\Xsf=x}} \mu_\Xsf(\dv x)
    \\& =\inf_{f\in\measf(\Omega, S)} \int
	    \inp{f(x), \mu_{\Ysf|\Xsf=x}} \mu_\Xsf(\dv x)
    \\& =\inf_{f\in\measf(\Omega, S + \R+^{n})} \int\inp{f(x), 
	      \mu_{\Ysf|\Xsf=x}} \mu_\Xsf(\dv x)
    \\&\overseteqref{eq:super_sum}{=}\inf_{f\in\measf(\Omega, \super(\ell))}
        \int\inp{f(x), \mu_{\Ysf|\Xsf=x}} \mu_\Xsf(\dv x)
    \\& =\inf_{f\in\measf(\Omega, \super(\ell))} \int f\dv \mu.\qedhere
  \end{align}
\end{proof}
\begin{proposition}\label{prop:dinf_support_representation}
  Suppose $E: [n] \expto \Omega$ is an experiment and $D\subseteq\R^n$ is
  closed. Then 
  \begin{gather}
    \I_{D}(E)= \sup_{f\in\measf(\Omega,D)}\int \sum_{i\in[n]} f_i \dv E_i.
    \label{eq:I-D-as-function-sup}
  \end{gather}
\end{proposition}
\begin{proof}
  The proposition follows from
  Lemma \ref{lem:integral_sprt_function_trick} with 
  $k(x,d) \defeq \inp{(\dv {E}/\dv \rho)(x),d}$ and 
  $\dv\rho \defeq \frac{1}{n}\sum_{i\in[n]} \dv E_i$,
  along with the observation that since $D$ is closed, it is also Borel
  measurable.
\end{proof}

Let $\pi\in\probm([n])$ be a prior distribution over $[n]$ together with a
binary experiment $E:[n]\expto\Omega$. Then there is a probability
distribution $\pi\times E\in\probm(\Omega\times[n])$ satisfying
\begin{gather}
	\Def{(\pi\times E)(\dv x, \dv y)}{\pi(\dv y)E(y,\dv x)}.
\end{gather}
It will be convenient to write the prior $\pi$ as a vector
$(\pi_1,\dots,\pi_n)\in\R^n$.

We now present the relationship between
$D$-information and the Bayes risk\footnote{
	This theorem (sans the geometric insight) was
  presented by \citet{Garcia-Garcia2012}, and restated in a related form
  by \citet{Duchi:2018wc}. It both extends and simplifies the version
  for $n = 2$ presented by \citet{Reid2011}, which itself extended
  beyond the symmetric (margin loss) case the version due to
  \citet{Nguyen2009}, which first appeared in \citep{Nguyen:2005},
  and which in turn extended the observations of
  \citet{OsterreicherVajda1993a} and \citep{Gutenbrunner:1990}.
  Earlier attempts to connect measures
  of information to Bayes risks include Fano's inequality
  \citep[Section 9.2]{Fano:1961uz}\citep[Section 5.3]{Polyanskiy:2019aa},
  and the inequalities derived by \citet{Perez:1967ud} and
  \citet{Toussaint:1974uc, Toussaint1977, toussaint78}.  Other
  precursors are the generalised entropies of \citet{Dupuis:2014tc}
  defined in terms of a Neyman-Pearson hypothesis testing problem (and thus
  equivalent to generalised variational divergence).  In the binary case,
  with Variational divergence and 0-1 loss, the bridge is classical
  \citep{Devroye:2013wq}.  There is now quite a literature  on
  information-theoretic statistical inference based on divergences
  \citep{Pardo:2018tp};   the bridge described in the present section
  suggests that such methods can be profitably viewed as a
  re-parametrisation of classical decision-theoretic methods based on
  expected losses.  The relationship between measures of information and
  the Bayes risk was also observed in \citep{chatzikokolakis2008bayes} for
  information security problems, and in \citep{Alvim2012measuring} for
  general information leakage problems.}.

\begin{theorem}\label{thm:bayes_risk_bridge}
  Suppose $\Omega$ is a standard Borel space,
  $\ell\in\measf(\probm([n])\times[n],\R)$, $E:[n]\expto\Omega$, and
  $\pi\in\probm([n])$.  Let  $\Def{\pi·\super(\ell)}{\setcond{(\pi\hadamard
  f)}{f\in \super(\ell)}}$ denote the Hadamard vector product $\pi\hadamard
  f=\rbr{\pi_1 f_1,\ldots,\pi_n f_n}$ for each element of $\super(\ell)$.
  Then
  \begin{gather}
    {\color{ForestGreen}
    \brisk_\ell(\pi\times E) = -\I_{-\pi \hadamard\super(\ell)}(E).
    \label{eq:bayes_risk_bridge}
    }
  \end{gather}
\end{theorem}
\begin{proof}
  Equation \eqref{eq:bayes_risk_bridge} is obtained from
  Lemma \ref{lem:brisk_with_superprediction_set} and
  Proposition \ref{prop:dinf_support_representation} as follows:
  \begin{align}
    \brisk_\ell(\pi\times E)
     & = \inf_{f\in\measf(\Omega, \probm([n]))} 
       \sum_{y\in[n]}\int \ell(f(x), y) \pi(y) E(y, \dv x)
     \\&
     \oversetref{L\ref{lem:brisk_with_superprediction_set}}{=}
     \inf_{f\in\measf(\Omega, \super(\ell))}\int\sum_{y\in[n]}f_y(x)
          \pi(y) E_y(\dv x)
    \\& = \inf_{f\in\measf(\Omega, \pi\hadamard\super(\ell))}\int\sum_{y\in[n]}f_y(x)
         E_y(\dv x).
    \\&\oversetref{P\ref{prop:dinf_support_representation}}{=}
    -\I_{-\pi·\super(\ell)}(E).\qedhere
  \end{align}
\end{proof}

\begin{remark}\label{rem:bridge-n-2}
	\normalfont
	The relationship $D=-\pi\hadamard\super(\ell)$ is a generalisation of that
	developed for $\phi$-divergences ($n=2$) in \citep{Reid2011} as we
	now elucidate. Inverting the relationship we have
	$\super(\ell)=-\left(\Tfrac{1}{\pi}\right) \hadamard D$, where
	$\Tfrac{1}{\pi}\defeq\left(\frac{1}{\pi_1},\ldots,\frac{1}{\pi_n}\right)$.
	It is elementary (and also follows using $r\hadamard
	x=\operatorname{diag}(r) x$ from 
	\citep[Proposition C.3.3.3]{hiriarturruty2001fca}) that
	\[
		\sigma_{-\left(\Tfrac{1}{\pi}\right)\hadamard
		D}(x)=\sigma_D\left(-\left(\Tfrac{1}{\pi}\right)\hadamard
		x\right).
	\]
	Setting $D=\hyp(-\phi^*)$ we know from Proposition 
	\ref{thm:fdiv_equivalence} that $\sigma_D=\breve\phi$ and thus 
	\begin{gather}
	\sigma_{\super(\ell)}(x)=
	\breve\phi\left(\frac{-x_1}{\pi_1},\frac{-x_2}{\pi_2}\right) 
	=\frac{-x_2}{\pi_2}\phi\left(\frac{x_1}{x_2}\frac{\pi_2}{\pi_1}\right).
	\label{eq:sigma-super-ell-phi}
	\end{gather}
	The negative support function of the superprediction set $\super(\ell)$
	corresponds to the conditional Bayes risk $\underline{L}^\pi$ 
	in \citep[Theorem 9]{Reid2011} (confer remark
	\ref{rem:bayes_risk_support_function}). 
	Parametrising in the same manner with
	$(x_1,x_2)=(\eta,1-\eta)$ and $(\pi_1,\pi_2)=(\pi,1-\pi)$ and
	substituting into \eqref{eq:sigma-super-ell-phi} we obtain
	\[
		\sigma_{\left(\frac{-1}{\pi},\frac{-1}{1-\pi}\right)\hadamard
		D}((\eta,1-\eta)) =	\underline{L}^\pi(\eta)=-\frac{1-\eta}{1-\pi}
		\phi\left(\frac{1-\pi}{\pi} \frac{\eta}{1-\eta}\right),
	\]
	consistent with \citep[Theorem 9]{Reid2011}. The loss 
	$\ell$ can be recovered from $\sigma_{\super(\ell)}$ via the
	derivative:
	$\ell=\operatorname{D}\sigma_{\super(\ell)}$
	\citep{williamson2014geometry, Williamson:2022vm}.  Evaluating the partial derivatives
	we obtain explicit formulae for $\ell_1$ and $\ell_2$ in terms of
	$\phi$:
	\begin{gather}
	\ell_1(x_1,x_2)=-\frac{1}{\pi_1}
		\phi'\left(\frac{x_1}{x_2}\frac{\pi_2}{\pi_1}\right) 
		\ \ \ \ \ \ \ \ \ \ \ 
	\ell_2(x_1,x_2)=
		 -\frac{1}{\pi_2}\phi\left(\frac{x_1}{x_2}\frac{\pi_2}{\pi_1}\right)
		 +\frac{1}{\pi_1}\frac{x_1}{x_2}\phi'\left(
		 \frac{x_1}{x_2} \frac{\pi_2}{\pi_1}\right),
	\end{gather}
	which are 0-homogeneous in $x$ as expected from Euler's homogeneous
	function theorem (see below).
\end{remark}

\subsection{The witness to the supremum in
	\texorpdfstring{$\I_D$}{D-information}}

$\I_D$ is defined via a supremum. There is insight to be
had by examining the function that attains this.  Let $\nabla\sprt_D$ be a
selection of $\subdiff\sprt_D$.  Euler's homogeneous function theorem:
 \begin{gather}
   \forall{x\in\dom \subdiff\sprt_D} \sprt_D(x)=\inp{\nabla\sprt_D(x), x},
   \label{eq:eulers_homog_theorem}
 \end{gather}
 and the 1-homogeneity of $\sprt_D$ implies $\subdiff\sprt_D$ is
 0-homogeneous  (so for any $c>0$, $\subdiff\sprt_D(cx)=\subdiff\sprt_D(x)$). 
 We can thus determine the argmax in \eqref{defn:d_information}:

 \begin{proposition}\label{prop:dinf_witness}
       Suppose $E: [n]\expto \Omega$ is an experiment and 
       $D\subset\R^n$.
       Let $\rho$ be a measure that dominates each of the measures $(E_y)_{y\in Y}$.
       Then if $\sprt_D$ is finite on $\R++^n$ there exists a selection
       $\nabla\sprt_D\in\subdiff\sprt_D$ over $\R++^n$, and
       \begin{align}
         \I_D(E)
         = \sup_{p\in\measf(X,\probm(Y))}
         \sum_{y\in[n]}\int_X(\nabla\sprt_D\circ p)_y \dv E_y
         = \sum_{y\in[n]}\int_X\rbr{\nabla\sprt_D\circ
           \diff{E}{\rho}}_y \dv E_y.    
       \end{align}
 \end{proposition}
 Note that the requirement is only that $\sprt_D$ is finite on $\R_{>0}^n$,
 not on $\R+^n$ which would exclude standard unbounded
 information measures such as Kullback-Leibler divergence.
 \begin{proof}
       Since $\R_{>0}^n\subseteq\relint(\dom \sprt_D)\subseteq \dom\subdiff\sprt_D$,
       by the Michael selection theorem
       \citep[Theorem 17.66, p.~589]{aliprantis2006infinite}
       there exists a continuous selection $\nabla\sprt_D$ mapping
       $\R_{>0}^n \to D$.  From the definition of the support function
       \begin{align}
         \sprt_D(x) = \sup_{d\in D} \inp{x,d} \geq
         \sup_{z\in\R++^n}\inp{x, \nabla\sprt_D(z)}.
       \end{align}
       By Euler's homogeneous function theorem, \eqref{eq:eulers_homog_theorem},
       $\sprt_D(x) = \inp{x, \nabla\sprt_D(x)}$, and consequently\!
       $\sprt_D(x)= \sup_{f\in\R^n_{> 0}}\inp{x, \nabla\sprt_D(f)}$ for all
       $x\in \R++^n$.
       Proposition \ref{prop:dinf_support_representation} implies that

       \begin{align}
         \I_D(E) & = \int \sprt_D\rbr{\diff{E}{\rho}} \dv\rho
         \\&= \int  \sup_{f\in\R++^n}\inp{\diff{E}{\rho}(x),
           \nabla\sprt_D(f)}\dv\rho(x)
         \\&= \int
         \sup_{d\in\nabla\sprt_D(\R++^n)}
         \inp{\diff{E}{\rho}(x), d}
         \dv\rho(x)
         \\&= \sup_{f\in\measf(\Omega, \nabla\sprt_D(\R++^n))}
         \sum_{y\in[n]}\int f_y \dv E_y
         \\&= \sup_{f\in\measf(\Omega, \nabla\sprt_D(\R++^n))}\sum_{y\in[n]}
         \int(\nabla\sprt_D\circ f)_y \dv E_y,
       \end{align}
       where in the fourth equality we apply Lemma
       \ref{lem:integral_sprt_function_trick} with $k = \inp{\marg,\marg}$.
       This proves the first equality since $\nabla\sprt_D$ is
       0-homogeneous.
    
       Euler's homogeneous function theorem implies
       \begin{align}
         \I_D(E)
          & =\int\sprt_D\rbr{\diff{E}{\rho}} \dv \rho
         \\& =
         \int \inp{ \diff{E}{\rho}, \nabla \sprt_D\circ
           \diff{E}{\rho} } \dv\rho
         \\&= \sum_{y\in [n]}\int\rbr{\nabla\sprt_D\circ \diff{E}{\rho}}_y \dv E_y.
       \end{align}
   This shows the second equality.
 \end{proof}

\begin{remark}
\label{rem:witness-phi}
\normalfont
    It is instructive to evaluate the witness
    of the supremum in Proposition \ref{prop:dinf_witness}
    in the case of
    $Y=\cbr{1,2}$ in terms of the $\phi$-divergence parameterisation 
    of $\I_D$.
    With $D_\phi$ as in \eqref{eq:D-phi-def},
    we have $\sigma_{D_\phi}=\breve{\phi}$. Assume
    $\phi$ is differentiable, so
    $\Def{\gain_\phi}{\Ds\sprt_{D_\phi}=\Ds\breve{\phi}}$ exists and 
    by direct calculation we obtain
    \begin{gather}
    	\gain_{\phi}(x,y)= \left(\begin{array}{c}
    		\phi'\left(\varfrac{x}{y}\right) \\ 
    		\phi\left(\varfrac{x}{y}\right)
    		-\phi'\left(\varfrac{x}{y}\right)\varfrac{x}{y}\end{array}\right),
	\label{eq:gain-from-phi}
\end{gather}
    with the witness is given by	
    \begin{gather}
    	\textstyle\gain_\phi\circ \frac{\dv E}{\dv\rho} =
    	\gain_\phi\rbr{\frac{\dv E_1}{\dv\rho},  \frac{\dv E_2}{\dv\rho}}
    	=\left(\begin{array}{c}
    		\phi'\left(\varfrac{\dv E_1}{\dv E_2}\right)\\
    		\phi\left(\varfrac{\dv E_1}{\dv E_2}\right) - 
    		\phi'\left(\varfrac{\dv E_1}{\dv E_2}\right)\varfrac{\dv E_1}{\dv E_2}
    	\end{array}\right)
\end{gather}
    and thus
    \begin{align}
    	\I_{D_\phi}(E)  & = \E_\rho\textstyle\left\langle \frac{\dv E}{\dv\rho},\,
    	   \gain_\phi\circ\frac{\dv E}{\dv\rho}\right\rangle\\
    	&= \int\textstyle\left[\frac{\dv E_1}{\dv\rho} 
		\phi'\left(\frac{\dv E_1}{\dv E_2}\right)
    		+\frac{\dv E_2}{\dv \rho}
		  \left(\phi\left(\frac{\dv E_1}{\dv E_2}\right) 
    		- \phi'\left(\frac{\dv E_1}{\dv E_2}\right)\frac{\dv E_1}{\dv
    		E_2}\right)\right]\dv\rho\\
    	& =\int\textstyle\phi'\left(\frac{\dv E_1}{\dv E_2}\right)\dv E_1 +
    	\displaystyle\int\textstyle\phi\left(\frac{\dv E_1}{\dv E_2}\right)
	\dv E_2 - \displaystyle\int\textstyle
    	\phi'\left(\frac{\dv E_1}{\dv E_2}\right) \dv E_1\\
    	&= \int\textstyle\phi\left(\frac{\dv E_1}{\dv E_2}\right) \dv E_2,
\end{align}
    which is the classical form of the $\phi$-divergence
    \eqref{def:fdiv_fmi}.
\end{remark}

\subsection{The Family of \texorpdfstring{$D$}{D}-informations}

\begin{proposition}\label{prop:d_inf_domain}
  Suppose $D\subseteq \R^n$ is convex. 
  Then $\tcl(\dom\sprt_D) = \R+^n$ if and only if $\rec(D) = \R-^n$.
\end{proposition}
\begin{proof}
  From \citep[Theorem 2.2.1~(c), p.~32]{Auslender2003}
  and the bipolar theorem we have
  \[
    \rbr{\dom\sprt_D}^* = \rec(D) = \R+^n
    \iff \tcl\rbr{\dom\sprt_D} = \rec(D)^* = \R-^n.\qedhere
  \]
\end{proof}

In Theorem \ref{thm:bayes_risk_bridge} we observed an interesting 
connection between the Bayes risks associated to a risk 
minimisation and the negative $D$-information associated with 
its negative superprediction set. There is also a similar 
asymptotic characterisation of the superprediction sets of 
positive loss functions.

Proposition \ref{prop:bayes_risk_domain} is a special case of a 
much more general result stated for superprediction sets on general 
outcome spaces in 
\citep[Proposition 4.6~(a), p.~67]{crankoAnalyticApproachStructure2021}.
We include its short proof for completeness.

\begin{proposition} \label{prop:bayes_risk_domain}
  Let $\ell:\probm([n])\to\R+^n$. Then $\rec(\super(\ell)) = \R+^n$.
\end{proposition}
\begin{proof}
  We first use the property that $A\subseteq B$ implies $\rec(A)
  \subseteq\rec(B)$ and $\ell(\probm([n]))\subseteq\R+^n$  to obtain
  \begin{gather}
    \rec\rbr{\super(\ell)} =
    \rec\rbr{\ell(\probm([n])) + \R+^n}
    \subseteq\rec\rbr{\R+^n + \R+^n} = \R+^n.
  \end{gather}
  This shows $\rec\rbr{\super(\ell)}\subseteq\R+^n$. Next, using the
  associativity of the Minkowski sum
  \begin{gather}
    \R+^n + \super(\ell) = \R+^n + \probm([n]) + \R+^n
    = \probm([n]) + \R+^n = \super(\ell),
  \end{gather}
  which shows $\R+^n\subseteq \rec\rbr{\super(\ell)}$, and completes the proof.
\end{proof}

After observing that $-\rec(D) = \rec(-D)$,
Propositions \ref{prop:d_inf_domain} and \ref{prop:bayes_risk_domain} yield
another characterisation of the connection between the $D$-information and
Bayes risks with nonnegative proper loss functions, this time in terms of
the asymptotic geometry of these sets.  Although it may seem coincidental
that---despite very different origins and motivating definitions---the sets
$\super(\ell)$ and $-D$ look very similar from afar, this relationship is
not at all surprising when parameterising these functionals using a set, as
we have done. The bilinearity of the expectation operator means that we are
working with a pointwise infimum or supremum over linear forms, that means
that, without loss of generality, we can replace the set by its closed
convex hull. This explains the natural characterisation in terms of
the support function (Remarks \ref{rem:d_inf_support_function} and
\ref{rem:bayes_risk_support_function}). 
Since both of these functionals operate on sets of probability measures, 
in order for them to be meaningful they should be sufficiently finite, this 
is the essence of the asymptotic characterisations in
Propositions \ref{prop:d_inf_domain} and \ref{prop:bayes_risk_domain}.

\begin{remark}
	\normalfont
We have shown that the recession cone of $\super(\ell)$ is such that the 
induced $D$ has the right recession cone for $D$ information, 
but what about normalisation? In the same way that there is some freedom in 
normalising $D$, we have freedom in normalising $\ell$.  
In previous work \citep{Vernet2016,williamson2014geometry, Williamson:2022vm} we have 
normalised proper losses $\ell$
such that $\ell(e_i)=0$ for $i\in[n]$ (where $e_i$ is the canonical 
unit vector). 
This implies that $\super(\ell)\subset\R_{\ge 0}^n$.  For the present paper it is 
more convenient to normalise such that 
\begin{gather}
    \label{eq:spr-normalisation}
    \ell(1_n/n)=0_n \mbox{\ \ and\ \ }\sigma_{\super(\ell)}(1_n/n)=0 .
\end{gather}
The first  condition implies that $0\in\tbd \super(\ell)$ and the second that
$\super(\ell)\subseteq\lev_{\le 0}\langle\cdot,-1_n\rangle$.  The bridge from
risks to information requires the specification of the prior $\pi$ which
can be seen to effectively scale $\super(\ell)$ separately in each dimension.
Of course the simplest case to consider is that $\pi=1_n/n$, in which case
it follows immediately that if $\ell$ satisfies
\eqref{eq:spr-normalisation} then $D\coloneqq -\pi\hadamard\super(\ell)$
satisfies $\sigma_D(1_n)=0\Rightarrow D\subseteq\lev_{\le 0} \langle \cdot,
1_n\rangle$ and $0\in\tbd D$, and consequently $D\in\Dfrak^n$.  Given an
$\ell$ that does not satisfy \eqref{eq:spr-normalisation}, it can be made
to do so by translation and scaling.  Thus the normalisation conditions we
impose upon $D$ can always be met by suitable adjustment of $\ell$.
Adopting the normalisation in \eqref{eq:spr-normalisation} means that for
$\pi=1_n/n$, the \emph{statistical information} of \citet{DeGroot1962}
is simply the negative Bayes risk, because $\sigma_{\super(\ell)}(1_n/n)=0$
implies the ``prior Bayes risk'' $\underline{\mathbb{L}}(\pi,M)$ is zero;
see \citep[Sections 4.6 and 4.7]{Reid2011}.
\end{remark}

\begin{remark}
	\normalfont
	The bridge result (Theorem \ref{thm:bayes_risk_bridge}) implies
	that any means by which multiple loss functions are combined, by
	combining their superprediction sets, provides an
	analogous combination scheme for information measures, by combining
	$D_i\in\Dfrak^n$, $i\in[m]$. The combination schemes in 
	\citep{Williamson:2022vm} based upon $M$-sums 
	\citep{Gardner:2013aa} suggest one can simply take $M$-sums of the
	$D_i$. This generalises the combination schemes proposed by
	\citet{Kus:2003aa,Kus:2008aa}.
\end{remark}


\subsection{\texorpdfstring{$D$}{D}-Information Processing Equality}

One of the most basic results in information theory is the information
processing inequality \citep{Cover:2012aa}.  It is often stated in terms of
mutual information, but there is a version, which is equivalent, in terms
of divergences \citep{Polyanskiy:2019aa}.  We defer until part II of the
paper \citep{Williamson:2023aa} a detailed statement and examination of the
connection between the two types, and indeed the connection with what we
present below.  Rather than an \emph{inequality}, below we present and
information processing \emph{equality}, with, however, a different measure
of information on either side of the equation.  Also, the result below is
for what in the machine learning community is called ``label noise''. The
traditional information processing inequality is for the situation of
``attribute noise'' and is treated in \S\ref{sec:Fscr-Nfrak-info} below.

\begin{proposition}\label{prop:label-noise}
  Suppose $E:[n]\expto \Omega$, $R: [n]\expto [n]$, and $D \subseteq \R^n$. Then
  \begin{gather}
	{\color{ForestGreen}
    \I_D(RE) = \I_{R^\trsp D}(E),\label{eq:IDSE}
    }
  \end{gather}
  where $R^\trsp D \defeq \setcond*{R^\trsp d}{d\in D}$ uses the representation of $R$ as a matrix.
\end{proposition}
\begin{proof}
  Identifying $R$ with its representation as a stochastic matrix, and writing
  $\varfrac{\dv  E}{\dv \rho}$ for the vector 
  $(\varfrac{\dv E_1}{\dv \rho},\dots, 
  \varfrac{\dv E_n}{\dv \rho})$ there is 
  $\varfrac{\dv{(RE)_i}}{\dv \rho} = (R \varfrac{\dv E}{\dv \rho})_i$ for
  $i\in[n]$, and

  \begin{align}
    \I_D(E)
     & =\int \sup_{d \in D}\rbr{ \sum_{i \in [n]} d_i\diff{(RE)_i}{\rho}(x)}  
       \rho(\dv x)
    \\
     & =\int \sup_{d \in D}\inp{d,R\diff{E}{\rho}(x)} \rho(\dv x)
    \\
     & =\int \sup_{d \in D}\inp{R^\trsp d,\diff{E}{\rho}(x)} \rho(\dv x)
    \\
     & =\int \sup_{d \in R^\trsp D}\rbr{ \sum_{i \in [n]} 
        d_i\diff{E_i}{\rho}(x)}\rho(\dv x)
    \\
     & = \I_{R^\trsp D}(E).\qedhere
  \end{align}
\end{proof}

\begin{remark}
	\normalfont
Observe that a $n\times n$ permutation matrix $P$ can be thought of as a
Markov kernel $P\colon[n]\expto [n]$.  Let $PD\coloneqq\{Pd\st d\in D\}$
and say that $D\subseteq\R^n$ is \Def{permutation
invariant} if for any such $P$, $PD=D$. Then Proposition
\ref{prop:label-noise} implies for such $D$ that $\I_D(PE)=\I_{P^\trsp
D}(E)=\I_D(E)$, since $P^\trsp$ is also a permutation matrix. Thus, in this
situation, $\I_D$ is permutation invariant.  An equivalent,
but less elegant, version of this observation was given in
\citep{Garcia-Garcia2012}.  When $n=2$, and $D$ is parametrised by $\phi$
as in Theorem \ref{thm:fdiv_equivalence}, this invariance corresponds to
the requirement that for all $x>0$, $\phi(x)= \phi^\diamond(x)=x\phi(1/x)$
(the Csiz\'{a}r conjugate of $\phi$) which implies
$\I_\phi(P,Q)=\I_\phi(Q,P)$.  
\end{remark}

\begin{example}[Label Noise]
	\label{ex:label-noise}
	\normalfont
  Let $E:[n]\expto\Omega$ be an experiment and $R: [n]\expto [n]$ a Markov
  kernel. We can thus form the product experiment $RE$ as per the diagram
  \begin{align}
    Y \mto{R} \tilde Y \mto{E} X.
  \end{align}
  This corresponds to ``label noise'' --- that is noise in the observations
  of $Y$. Instead of learning from $(X,Y)$ one only gets to observe
  $(X,\tilde{Y})$ for some corrupted version $\tilde{Y}$ of the true label
  $Y$. For example, when $n = 2$, and $Y\in[2]$ one might have a label flip
  with probability $p$. This corresponds to $R$ having the representation
  as the stochastic matrix
  \begin{gather}
    R\defeq
    \begin{pmatrix}
      1-p & p    \\
      p   & 1-p.
    \end{pmatrix}
  \end{gather}
  Then for $D\subseteq\R^n$, with Proposition \ref{prop:label-noise}
  \begin{gather}
    \I_{D}(RE) = \I_{R^\trsp D}(E).
  \end{gather}
\end{example}

When $n=2$ one can translate the result of Proposition
\ref{prop:label-noise} to the language of $\phi$ divergences, in which form
the result is less perspicuous than  \eqref{eq:IDSE}:
\begin{figure}[t]
	\begin{center}
	\includegraphics[width=0.7\textwidth]{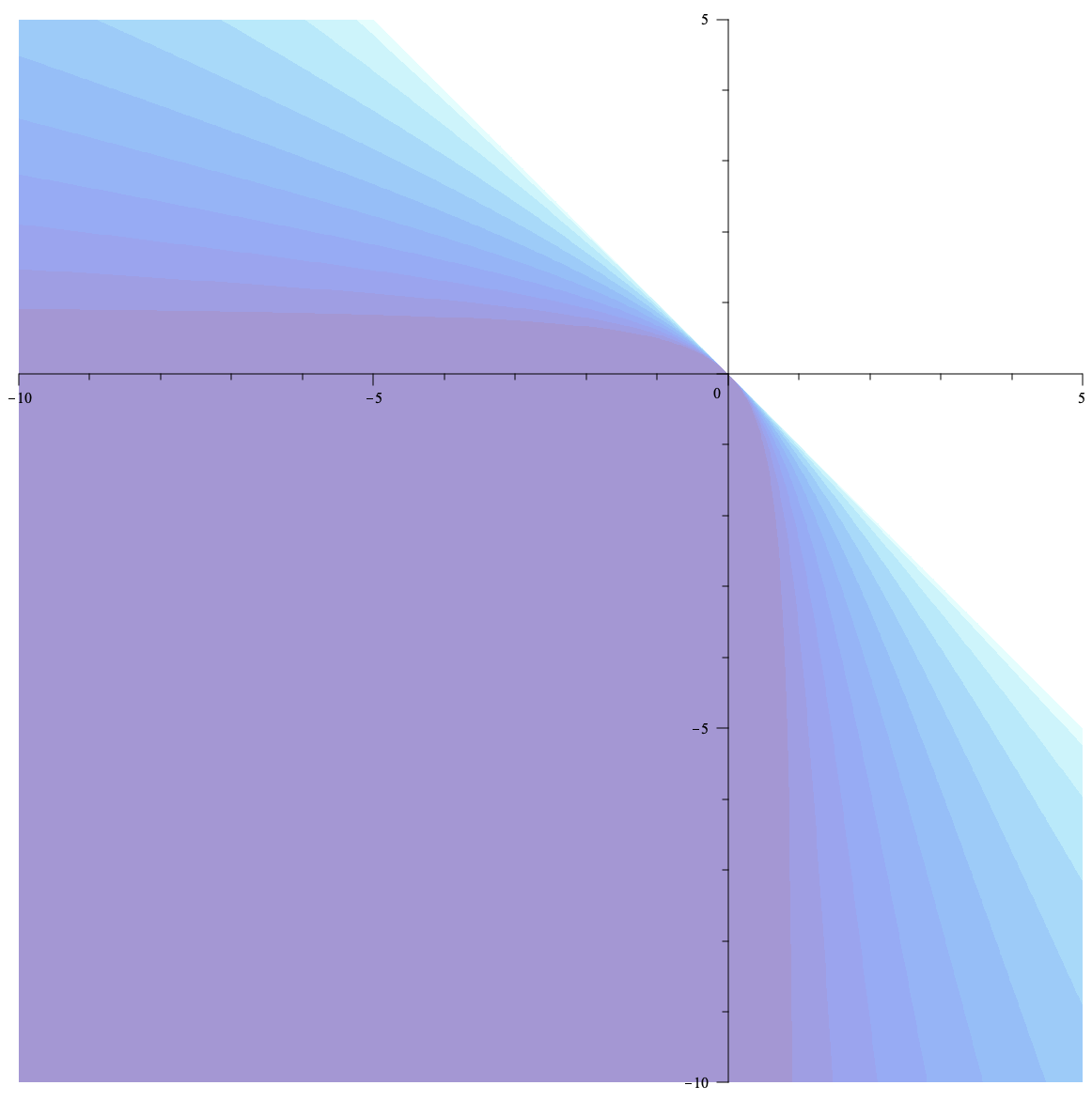}
	\end{center}

	\caption{Effect of $R_r$ on $D_{\mathrm{Hell}}$.
		The  plot
		shows $R_r^* D_{\mathrm{Hell}}$ (restricted to $[-10,5]^2$)
		for $r$ ranging from 0.5
		(very pale blue) to $1.0$ (purple), which is of course just
		$D_{\mathrm{Hell}}$.  \label{fig:C-D-r}}
\end{figure}
\begin{corollary} \label{corr:phi-div-R-channel}
	Suppose  $\phi\in\Phi$, 
	$R\colon[2]\expto [2]$ is the Markov Kernel parametrised as
	\begin{gather}
	\cramped{R=\left[\begin{array}{cc} r_1 & 1-r_1\\ 1-r_2 &
			r_2\end{array}\right]},
		\mbox{\ \ \ and\ \ \ }
		\phi_R(z)= \left((1-r_2)z +r_2\right) 
		\phi\left(\frac{r_1 z+1-r_1}{(1-r_2)z+r_2}\right).
	\end{gather}
	Then for all experiments $E\colon [2]\expto \Omega$,
	$\I_\phi(RE)=\I_{\phi_R}(E)$.
\end{corollary}
\begin{proof}
	The trick is to identify the perspective of
	$\phi$ with the support function of $D$:
	$\breve{\phi}(x,y)=\sprt_D( (x,y)^\trsp)$ for $x,y,\in\R_{\ge 0}$.
	The effect of the Markov kernel $R$ on $\phi$ can be determined from
	\[
		\breve{\phi}_R(x,y) =\sprt_D(R\cdot(x,y)^\trsp)=\breve{\phi}(R\cdot
		(x,y)^\trsp) = \breve{\phi}(r_1 x + (1-r_1)y, (1-r_2)x + r_2 y).
	\]
	Consequently $\I_\phi(RE)=\I_{\phi_R}(E)$,
	where
	\[
	 \phi_R(z)  = \breve{\phi}_R(z,1)
		 = \breve{\phi}\left(r_1 z +1-r_1, (1-r_2)z+r_2\right)
		= \left((1-r_2)z +r_2\right) \phi\left(\textstyle\frac{r_1
		z+1-r_1}{(1-r_2)z+r_2}\right).\qedhere
	\]
\end{proof}

\begin{remark}
	\normalfont
	Example \ref{ex:label-noise} corresponds to previous work on loss
	correction, whereby learning with a given loss with noisy labels is
	equivalent to learning with a ``corrected loss'' with noiseless
	labels; see e.g.~\citep{van2015learning, Patrini:2017aa,
	Rooyen:2018aa}.
\end{remark}

Figure \ref{fig:C-D-r} illustrates this case for the binary
\emph{symmetric} channel
$R_r=\scriptsize\left[\begin{array}{cc} 
		r & 1-r\\ 
		1-r & r
	\end{array}\right]$ 
for $D_{\mathrm{Hell}}$ (corresponding to squared Hellinger divergence ---
see table~\ref{table:phi-divergences} in Appendix \ref{sec:var-rep}).
Observe that (as motivated in
Appendix \ref{sec:expected-gauge}) if $D^\polar=C$, then $(R^*D)^\polar=
R^{-1} C$ (this follows simply by substitution into the definition of the
polar).

\section{Constrained Information Measures --- $\mathcal{F}$-information}
\label{sec:Fscr-Nfrak-info}
As we saw in \S\ref{subsec:bridge}, Theorem \ref{thm:bayes_risk_bridge} 
shows how the $D$-information can be connected to risk minimisation. 
In practice one can never actually attain the Bayes risk, because with access 
only to a finite number samples rather than the exact underlying 
distribution, one needs to restrict the hypothesis class 
\citep{Vapnik1998} in order to make the optimisation 
in \eqref{defn:bayes_risk} well-posed (when using empirical measures).
It is helpful to consider the formula for the (unconstrained) Bayes risk
in 
\eqref{defn:bayes_risk} repeated below for convenience.
\begin{gather}
\brisk_\ell(\mu)
      =\inf_{f\in\measf(\Omega, \probm([n]))} \E_{(X,Y)\sim\mu}\sbr{\ell(f(X), Y)}.
\end{gather}
Embracing the above viewpoint, 
we modify this by optimising over
$\cal{H}\subsetneq \measf(\Omega,\probm([n]))$, and call this restriction the
\Def{constrained Bayes risk}:
\begin{gather}
	\Def{\brisk_{\ell, \cal H}(\mu)}{
		\inf_{h\in\cal H}\E_{(X,Y)\sim \mu}\sbr{\ell(h(X), Y)}}.
  \label{defn:constrained_bayes_risk}
\end{gather}
There is a slight redundancy in \eqref{defn:constrained_bayes_risk} since the
function class $\cal H$ only appears via composition with the loss 
function $\ell$. When viewed in terms of information rather than 
risk, \eqref{defn:constrained_bayes_risk} is precisely the measure 
of information which we now introduce.

\subsection{\texorpdfstring{$\mathcal F$}{F}-information}

Recall the expression for $\I_D(E)$ in \eqref{eq:I-D-as-function-sup}
(swapping the integral and the sum for convenience in what follows):
  \begin{gather}
    \I_{D}(E)= \sup_{f\in\measf(\Omega,D)} \sum_{i\in[n]}\int f_i \dv E_i.
  \end{gather}
If now we restrict the supremum to be over 
a set $\mathcal F \subseteq \measf(\Omega,D)$, 
the \Def{$\cal F$-information} of an experiment $E:[n]\expto\Omega$ 
is\footnote{There are a number of precursors of $\I_{\cal F}$ 
   which are summarised in Appendix \ref{sec:precursors}.}
\begin{gather}
	\Def{\I_{\mathcal F}(E)}{ \sup_{f\in \mathcal F} \sum_{i\in[n]}\int
	f_i\dv E_i}.\label{def:f_information}
\end{gather}
With notation that is consistent with Theorem \ref{thm:bayes_risk_bridge},
from the definition of the constrained Bayes risk
\eqref{defn:constrained_bayes_risk}, a prior $\pi\in\probm([n])$, a loss
function $\ell$, an experiment $E$, and a hypothesis class $\cal H
\subseteq \measf(\Omega,\probm([n]))$, the $\cal F$-information is related
to the constrained Bayes risk by
\begin{gather}
  \brisk_{\ell, \cal H}(\pi \times E)
  = -\I_{-\pi·\ell \circ \cal H}(E),
\end{gather}
where $\pi·\ell \circ \cal H\defeq \setcond{x\mapsto(\pi_1\ell(h(x), 1),
\dots, \pi_n\ell(h(x), n)) }{ h \in \cal H}$, that is the composition 
of $\cal H$ with $\ell$, and scaled by $\pi$. 
(This is proved below in Theorem \ref{thm:fscr_brisk_bridge}.)

 Consider the collection $\measf(X,D)$ of measurable mappings from $X$ to
 some $D\in\cvxrec(\R^n,\R-^n)$.   For $\cramped D\in\cvxrec(\R^n,\R-^n)$
 denote by $ \Def{\scr{C}_D}{\setcond{x\mapsto d}{ d\in D}} $ the \Def{set of
 constant maps from $X$ to $D$}.
 We say that $\mathcal{F}$ is \Def{$D$-ranged} if
 $\mathcal{C}_D\subseteq\mathcal{F}\subseteq\mathcal{L}_0(X,D)$.  If
 $\mathcal{F}\subseteq\mathcal{L}_0(X,\R^n)$,  let
 \begin{gather}
   \Def{{\mathcal F}(X)}{\bigcup_{f\in{\mathcal F}} f(X)},
 \end{gather}
 where $\Def{f(X)}{\setcond{f(x)}{ x\in X}}$. 
%
 If $D \in\cvxrec(\R^n,\R-^n)$  and
 $\mathcal{F}$ is $D$-ranged, then $\scr{F}(X)=D$ since 1) 
 $\scr{C}_D \subseteq\scr{F}\Rightarrow  \scr{F}(X)\supseteq D$ and 
 2) $\scr{F}\subseteq \measf(X,D) \Rightarrow
   \scr{F}(X)\subseteq D$.

 Every $D$-ranged $\mathcal F $ is a collection of appropriately measurable
 mappings from $X$  into $D$, and every $d\in D$ may be attained by $f(x)$
 for some $f\in\mathcal F$ and $x\in X$. The maximal (by subset
 ordering)  $D$-ranged $\mathcal{F}$ is simply $\measf(X,D)$, the set of
 \emph{all} measurable mappings from $X$ to $D$.  Choosing smaller sets is
 equivalent to working with restricted hypothesis classes in normal
 statistical decision problems (an assertion we make precise below).  The
 extra flexibility of working with such constrained function classes is
 necessary to capture the effects of attribute noise on information
 measures.

 Suppose  $D\in\cvxrec(\R^n,\R-^n)$,  $\mathcal{F}$ is $D$-ranged
 and $E: [n] \expto X$.
 The \Def{$\mathcal F$-information of $E$} is
 \begin{align}
   \Def{\I_{\mathcal F}(E)}& \defeq \Def{ \sup_{f\in\mathcal F} \sum_{y \in [n]}
     \int f_y \dv E_y} \label{eq:I-Fscr-def}\\
   &= \sup_{f\in\scr{F}}\sum_{y\in[n]} \int f_y \frac{\dv E_y}{\dv
     \rho} \dv\rho\\
   &=\sup_{f\in\scr{F}}\int\sum_{y\in[n]} f_y\frac{\dv E_y}{\dv
	   \rho}\dv\rho\\
	   & = \Def{\sup_{f\in\scr{F}}\E_\rho \inp{ \frac{\dv E}{\dv\rho}, f}},
      \label{eq:I-F-via-rho}
 \end{align}
 where $\rho$ is an arbitrarily chosen reference measure.  If
 $D\in\cvxrec(\R^n,\R-^n)$ and $\Def{{\mathcal F}_D}{ \measf(X,D)}$, then
 it is apparent from \eqref{defn:d_information} that $\I_{{\mathcal F}_D}
 = \I_{D}$; for any other ${\mathcal F}\subseteq \measf(X,D)$ we obviously
 have $\I_{\mathcal F}\le \I_D$, since the supremum is further restricted.
 If furthermore $D$ is normalised (i.e.~$D\in\Dfrak^n$) and
 $\mathcal{F}$ is $D$-ranged, then $0\le \I_\mathcal{F}(E)$ as can seen by
 considering $\scr{F}=\scr{C}_D$ whence
 \[
   \I_{\scr{F}}(E)=\sup_{d\in D}\sum_{y\in[n]} \int d\, \dv E_y=\sup_{d\in D}
   \sum_{y\in[n]} d=\sup_{d\in D} \inp{ d,1_n} =\sprt_D(1_n) =0.
 \]
 Thus for $D\in\Dfrak^n$
 and $D$-ranged $\scr{F}$,
 \begin{gather}
   0\le \I_{\scr{F}}(E)\le \I_D(E).
 \end{gather}
 

 The $\scr{F}$-information is invariant under convex hulls and closure
 \citep{Muller:1997aa}:
 \begin{proposition}\label{prop:cl-co-F}
   Suppose 
   $D\subset\R^n$,
   $E: Y\expto X$, $\I_D(E)<\infty$
   and $\scr{F}$ is $D$-ranged. Then
   \begin{gather}
     \I_{\scr{F}}(E)=\I_{\co\scr{F}}(E)=\I_{\clco\scr{F}}(E).
   \end{gather}
 \end{proposition}
 \begin{proof}
   Let $\Delta=\cbr{(\alpha_i)_i\st 0\le \alpha_i \ \forall
       i\mbox{\ and\ } \sum_i\alpha_i=1}$.  
   The convex hull of $\mathcal{F}$ is
   \[
	   \co\scr{F}=\setcond{\textstyle\sum_{i\in\mathbb{N}}
           \alpha_i f^i }{ f^i\in\scr{F},\
       \forall i\in\mathbb{N}
       \mbox{\ and\ } \alpha=(\alpha_i)_i\in\Delta}.
   \]
   Hence
   \begin{align}
     \I_{\co\scr{F}}(E) & = \sup_{\substack{f=\sum_i\alpha_i f^i\\
       f^i\in\scr{F},\alpha\in\Delta}}\ \sum_{y\in[n]} \int
     f_y \dv E_y                                                       \\
                        & =\sup_{\alpha\in\Delta} \sup_{f^i\in\scr{F}}
     \sum_{y\in[n]} \int \sum_i \alpha_i
     f_y^i \dv E_y\label{eq:cl-co-F-sum-swap1}                         \\
                        & = \sup_{\alpha\in\Delta}\sup_{f^i\in\scr{F}}
     \sum_i \alpha_i \sum_{y\in[n]} \int f_y^i \dv E_y
     \label{eq:cl-co-F-sum-swap2}                                      \\
                        & = \sup_{\alpha\in\Delta} \sum_i\alpha_i
     \sup_{f^i\in\scr{F}} \sum_{y\in[n]} f_y^i \dv E_y                  \\
                        & = \sup_{\alpha\in\Delta} \sum_i \alpha_i
     \I_{\scr{F}}(E)                                                   \\
                        & = \I_{\scr{F}}(E).
   \end{align}
   We need to justify the interchange of order of summation at
   \eqref{eq:cl-co-F-sum-swap1}--\eqref{eq:cl-co-F-sum-swap2}. The
   reordering can only fail if there are two subsequences, one
   diverging to $-\infty$ and one to $+\infty$ which cancel each
   other out. But this is impossible because
   $\I_{\scr{F}}(E)\le\I_D(E)<\infty$ and thus there can be no terms
   that diverge to $+\infty$ (even though it is possible that
   $f_i(x)=-\infty$, but such $f$ would not be chosen by the supremum
   operation, and all the $\alpha_i\in[0,1]$).
   This proves the  first equality.

   We can now assume $\scr{F}$ is convex. We need to show
   $\I_{\cl\scr{F}}(E)=\I_{\scr{F}}(E)$. But
   $\I_{\scr{F}}(E)=\sup_{f\in\scr{F}} \Psi(f,E)$, where $\Psi(f,E)
   =\sum_{y\in[n]} \int f_y \dv E_y$ is bounded since
   $\I_{\scr{F}}(E)\le\I_D(E)<\infty$. The function $f\mapsto \Psi(f,E)$ is
   also linear and thus continuous for any $E$. The supremum of a
   continuous real-valued function over the closure of a set is equal to
   the supremum over the set, which proves the second equality.
 \end{proof}

 Thus there is no loss of generality in henceforth assuming that ${\mathcal
 F}$ is closed and convex, as has been observed in the special case when
 $n=2$ and $D=D_\mathrm{var}$ (defined in Lemma \ref{lemma:Dvar-explicit})
 corresponding to ``integral probability metrics'' which are variants of
 variational divergence with a restricted function class
 \citep{Muller:1997aa}.  Equivalently, if ${\mathcal F}$ was not closed
 and convex, one can take the closed convex hull and not change the value
 of $\I_{\mathcal F}(E)$ (nor indeed change the  Rademacher complexity
 of $\mathcal{F}$ \citep{BartlettMendelson2002}). Since 
 convex function classes enable fast rates of convergence
 \citep{van-Erven:2015aa,MendelsonWilliamson2002} and optimization is in
 principle simpler, this is an appealing restriction, and one which is
 receiving practical attention in the form of infinitely wide neural
 networks \citep{Ergen2021}.  If $\mathcal{F}$ is closed and convex then so
 is $\mathcal{F}(X)$.

\subsection{The Bridge between \texorpdfstring{$\mathcal{F}$}{F}-Information 
and Constrained Bayes Risk}

 We now relate $\I_{\mathcal{F}}$ to the constrained Bayes risk
 \eqref{defn:constrained_bayes_risk} and to $\I_D$ \eqref{defn:d_information}:

 \begin{theorem}\label{thm:fscr_brisk_bridge}
   Suppose $\ell$ is a continuous proper loss, $E: [n]\expto X$ an experiment,
   $\pi\in\probm(Y)$ a prior distribution, and $\mathcal
     H\subseteq\measf(X,\probm([n]))$ an hypothesis
     class.  Let $\Def{\mathcal F}{\co(-\pi\hadamard \ell\circ\scr H)}$. 
   Then
   \begin{gather}
	{\color{ForestGreen}
     \brisk_{\ell\circ\scr H}(\pi\times E)  = -\I_{\mathcal F}(E).
     }
   \end{gather}
   Furthermore  
   $\scr F\subseteq\mathcal{L}_0(X,D)$, where $D=-\pi\hadamard \super \ell$.
 \end{theorem}
 \begin{proof}
   From the definition of the Bayes risk
   \begin{align}
     \brisk_{\ell\circ\scr H}(\pi\times E)
      & = \inf_{f\in\ell\circ\scr H}\sum_{y\in[n]} \int  \pi_yf_y  \dv E_y
     \\&= -\sup_{f\in\ell\circ\scr H}\sum_{y\in[n]} \int  -\pi_yf_y  \dv E_y
     \\&= -\sup_{-\pi \hadamard f\in\ell\circ\scr H}\sum_{y\in[n]} \int  f_y  \dv E_y
     \\&= -\sup_{f\in -\pi\hadamard \ell\circ\scr H}\sum_{y\in[n]} \int  f_y  \dv E_y
     \\&= -\I_{-\pi\hadamard \ell\circ\scr H}(E)\label{eq:bridge-sans-co}
     \\&= -\I_{\mathcal F}(E).\label{eq:bridge-with-co}
   \end{align}
 Whether one prefers the result \eqref{eq:bridge-sans-co} or
 \eqref{eq:bridge-with-co} is a matter of taste; the equality of 
 the two follows from Proposition \ref{prop:cl-co-F}.
     For the second part, we have $\ell\circ{\scr H}\subset \super\ell$. Thus 
     $\co(\ell\circ{\scr H}(X))\subset \co\super\ell=\super\ell$, 
     since $\super\ell$ is convex. Consequently,
     ${\scr F}(X)=\co(-\pi\hadamard\ell\circ{\scr H}(X)) =
     -\pi\hadamard\co(\ell\circ{\scr H}(X)) \subset 
     -\pi\hadamard\super\ell= D$, and thus ${\scr F}\subseteq\mathcal{L}_0(X,D)$.
 \end{proof}
\begin{remark}
	\normalfont
Observe that convexity of $\scr H$ does \emph{not} imply convexity of
$\ell\circ {\scr H}(X)$, where ${\scr H}(X)=\{h(x)\st h\in {\scr H},\ x\in
X\}$, but since $\super\ell$ is convex, we do have that for $h_0,h_1\in\scr
H$ and $h_\alpha=(1-\alpha)h_0+\alpha h_1$ that $\ell\circ
h_\alpha\in\super\ell$ for all $\alpha\in[0,1]$, and thus
$\I_{-\pi\hadamard\ell\circ\co{\scr H}}(E)=\I_{-\pi\hadamard\ell\circ{\scr
H}}(E)$, and so convexity of $\scr H$ does not ``hurt.''
\end{remark}

\subsection{\texorpdfstring{$\mathcal F$}{F}-Information Processing Equalities}

The  definition of $\cal F$-information \eqref{def:f_information},
implies an immediate, similar result to Proposition \ref{prop:label-noise}.
\begin{proposition}\label{prop:label-noise2}
  Suppose $E:[n]\expto \Omega$,
  $R: [n]\expto [n]$, and $\cal F \subseteq \measf(\Omega,\R^n)$. Then
  \begin{gather}
    \I_{\cal F}(RE) = \I_{R^\trsp\cal F}(E),     \label{eq:IFSE}
  \end{gather}
  where $R^\trsp \cal F \defeq \setcond{ x\mapsto R^\trsp 
  (f_1(x),\dots,f_n(x))^\trsp }{ f\in\cal F}$. Moreover, 
  if $D\subseteq \R^n$ and $\mathcal F\subseteq \measf(\Omega,D)$, 
  then $R^\trsp \mathcal F \subseteq \measf(\Omega,R^\trsp D)$.
\end{proposition}

The additional generality of $\cal F$-information yields another kind of 
information processing equality, one which is more aligned with the 
traditional formulation of information processing inequalities.
Rather than the ``processing'' being on the labels (the $Y$ in usual terminology) 
as in Proposition \ref{prop:label-noise2}: 
\[
    [n]\stackrel{R}{\expto} [n] \stackrel{E}{\expto}\Omega, 
\]
it is applied to the output (the $X$ of the experiment):
\[
    [n] \stackrel{E}{\expto} \Omega \stackrel{S}\expto \Omega .
\]
  For $f\in\mathcal{F}$, let $\Def{Sf}{(S
  f_1,\ldots, S f_n)}$, with $f=(f_1,\ldots,f_n)$ (i.e.~$f_i$ are the
  partial functions of $f$), and let  $\Def{S\cal F}{
  \setcond{Sf}{f\in\mathcal F}}$ (that is, the application of $S$
  is component-wise and element-wise).
\begin{theorem}\label{prop:I-F-ET}
  Suppose $E:[n]\expto \Omega$, $S: \Omega\expto \Omega$, and $\mathcal F
  \subseteq \measf(\Omega,\R+^n)$.  
  Then
  \begin{gather}
	  	{\color{ForestGreen}
    \I_{\cal F}(ES)=\I_{S\cal F}(E).
    }
  \end{gather}

\end{theorem}
\begin{proof}
  From the linearity of the integral, for each $i\in[n]$, and all $f\in\cal F$
  \begin{align}
    \int f_i(x)ES(i, \dv x)
     & = \int  f_i(x)\int E(i, \dv x')S(x', \dv x)
    \\& = \iint  f_i(x)S(x', \dv x)E(i, \dv x')
    \\&\overseteqref{def:markov_operators}{=}
    \int  (Sf_i)(x')E(i, \dv x').\label{eq:interchange}
  \end{align}
  In the final equality we apply Tonelli's theorem (Fubini's theorem for 
  sign-definite integrands) to exchange the 
  order of integration. We can do so since by assumption, 
  all $f\in\mathcal{F}$ are non-positive, and for all $x'$ and $i$ the 
  measures $S(x',\cdot)$ and $E(i,\cdot)$ are probability measures 
  and thus $\sigma$-finite.
  Thus
  \begin{align}
    \I_{\cal F}(ES)
     & =\sup_{f\in\cal F}\sum_{i\in[n]}\int f_i(x)(ES)_i(\dv x)
    \\&\overseteqref{eq:interchange}{=}
      \sup_{f\in S\cal F}\sum_{i\in[n]}\int f_i \dv E_i.
    \\&=\I_{S\cal F}(E).\qedhere
  \end{align}
\end{proof}

\begin{corollary}\label{thm:e_r_t_theorem}
  Suppose $E:[n]\expto \Omega$,
  $R: [n]\expto [n]$, and
  $S: \Omega\expto \Omega$, and $\mathcal F$ satisfies the conditions of 
  Theorem \ref{prop:I-F-ET}.
  Then
  \begin{gather}
    \I_{\cal F}(RES)=\I_{R^\trsp S\cal F}(E)=\I_{S R^\trsp \cal F}(E).\label{eq:rsf}
  \end{gather}
\end{corollary}
\begin{proof}
  Equation \eqref{eq:rsf} is obtained by applying Proposition 
  \ref{prop:label-noise2} and Theorem \ref{prop:I-F-ET}.
  The commutation of $R$ and $S$ is verified by applying Proposition
  \ref{prop:label-noise2} and Theorem \ref{prop:I-F-ET} in alternating orders.
\end{proof}

\begin{remark}
	\normalfont
  More generally, the condition in Theorem
  \ref{prop:I-F-ET} that $\mathcal F\subseteq\measf(\Omega,\R+)$ can be relaxed whenever there is $(ES)f = E(Sf)$. For example:
  \begin{enumerate}
    \item $\mathcal F \subseteq \measf(\Omega,\R-^n)$,
          using Tonelli's theorem with $-f$.
    \item $(\Omega,\lambda)$
          is a sigma-finite measure space,
	  $S(x, \marg) \ll \lambda$ is a Markov kernel, and $\mathcal
	  F\subseteq\lebf_1(\Omega,\R^n)$. Then there is a measurable
	  $k\in\measf(\Omega_1\times\Omega_2,\R+)$ with $k(x,y)\lambda(\dv
	  y) = S(x,\dv y)$, and
          \begin{align}
            \forall{\mu\in\probm(\Omega)}\iint \abs{f(x)k(x,y)}\mu(\dv x)
	       \lambda(\dv y) < \infty,
          \end{align}
          and using Lemma \ref{lem:density_interchange} below 
          we can apply Fubini's theorem.
    \item $\mathcal F \subseteq \contfb(\Omega,\R^n)$ (bounded continuous
	    functions), then we have the
	    dual pair $\inp{\contfb(\Omega), \signm(\Omega)}$
	    ($\signm(\Omega)$ is the space of finitely additive signed
	    measures on $\Omega$)  for which the
	    interchange \eqref{eq:interchange} is equivalent to the
	    existence of an adjoint of the linear operator $\mu\mapsto \mu
	    S$.
  \end{enumerate}
\end{remark}

\begin{lemma}\label{lem:density_interchange}
  Suppose $(\Omega_2,\lambda)$ is a measure space,
  $S:\Omega_1\expto\Omega_2$ is a Markov kernel, there is a measurable
  $k\in\measf(\Omega_1\times\Omega_2,\R+)$ with $k(x,y)\lambda(\dv y) =
  S(x,\dv y)$. Then for $f\in\lebf_1(\Omega_2,\R)$
  \begin{align}
    \forall{\mu\in\probm(\Omega_2)}\iint 
        \abs{f(y)k(x,y)}\lambda(\dv y)\mu(\dv x) < \infty.
  \end{align}
\end{lemma}
\begin{proof}
  Fix $\mu\in\probm(\Omega_2)$. Then
  \begin{align}
    \MoveEqLeft\iint \abs{f(y)k(x,y)}\mu(\dv x)\lambda(\dv y)
    \\& ≤
    \underbrace{\rbr{\iint\abs{f(y)}\mu(\dv x)\lambda(\dv y)}}_{\norm{f}}
    \underbrace{\rbr{\iint k(x,y)\mu(\dv x)\lambda(\dv y)}}_1.
  \end{align}
  The term in the second underbrace is $1$ because, observing the integrand
  is nonnegative and $S$ is a Markov kernel, we can apply Tonelli's theorem
  to obtain $\int k(x,y)\mu(\dv x)\lambda(\dv y) =  \int \mu(\dv x) = 1$.
  By hypothesis $\norm{f} < \infty$, which completes the proof.
\end{proof}

\subsection{The Information Processing Equality
   in terms of Constrained Bayes Risk}
\label{ex:attribute-noise}
  Let $E:[n]\expto\Omega$ be an experiment and  $S: \Omega\expto \Omega$ be
  Markov kernel.  We can thus form the product experiment $ES$ as per the
  diagram $ [n] \stackrel{E}{\expto} \Omega \stackrel{S}\expto \Omega $.
  This corresponds to ``attribute noise'' --- that is noise in the
  observations of $\Xsf$. For example, instead of learning from $(\Xsf,\Ysf)$ one
  only gets to observe $(\Xsf+\Nsf,\Ysf)$ for some independent noise random variable
  $\Nsf$. More general (non-additive) corruptions are possible, but this
  additive one will be of particular interest. 
  
  We can express the information processing equality in terms of Bayes risks:
  \begin{corollary}
      Suppose $\ell\colon\Delta^n\rightarrow\R_{\ge 0}^n$ is a
      continuous proper loss,  $\pi\in\probm([n])$ a prior distribution on $[n]$,
      and $\mathcal H\subseteq\measf(X,\probm([n]))$ an hypothesis
      class. Then
       \begin{gather}
         \brisk_{\ell\circ\scr H}(\pi\times ES) = \brisk_{S(\ell\circ\scr
	 H)}(\pi\times E).
       \end{gather}
  \end{corollary}
  \begin{proof}
   Let $\mathcal F = \co(-\pi\hadamard \ell\circ\scr H)$.
  By combining Theorem \ref{thm:fscr_brisk_bridge} with Theorem
  \ref{prop:I-F-ET} we have
   \begin{gather}
     \brisk_{\ell\circ\scr H}(\pi\times ES)  = -\I_{\mathcal F}(ES) = -\I_{S\cal F}(E)=
      \brisk_{S(\ell\circ\scr H)}(\pi\times E).
   \end{gather}
   The last equality is justified as follows.  
   For any set $A\subset\R^n$ and $S\colon [n]\expto [n]$, 
   we have $S\co A= \co (SA)$ since
   \begin{align}
       S \co A & = S\left\{\textstyle\sum_i \alpha_i a_i \st a_i\in A,\ 
             \alpha_i\ge0,\ \sum_i\alpha_i=1\right\}\\
             &=\left\{\textstyle\sum_i \alpha_i S a_i \st a_i\in A,\  
             \alpha_i\ge0,\ \sum_i\alpha_i=1\right\}\\
           &=\left\{\textstyle\sum_i \alpha_i b_i \st b_i\in SA,\  
             \alpha_i\ge0,\ \sum_i\alpha_i=1\right\}\\
             &= \co(SA).
   \end{align}
   Furthermore, for any $v\in\R^n$, any set $C\subset\R^n$ 
   and $S\colon [n]\expto [n]$, 
   we have $S(v \hadamard C)= v\hadamard SC$ since
   $
        S(v\hadamard C) = S\{v\hadamard c\st c\in C\}
        = \{(S v) \hadamard c\st c\in C\}
        = \{v \hadamard (Sc)\st c\in C\}
        = \{v \hadamard b\st b\in SC\}
        = v\hadamard SC.
   $
   These two facts together imply
   $S\cal F= S\co(-\pi\hadamard \ell\circ\scr H)=
   \co(S(-\pi\hadamard \ell\circ\scr H))=
   \co(-\pi\hadamard  S(\ell\circ\scr H))$, and a 
   second appeal to Theorem \ref{thm:fscr_brisk_bridge} concludes the proof.
   \end{proof}
  
\begin{remark}
	\normalfont

	Kernel methods in machine learning \citep{Scholkopf:2001wx} are so
	named because of the kernel of the integral operator $T_k\colon
	L_2\rightarrow L_2$ given by
	\begin{gather}
		\label{eq:integral-operator}
		T_k f = \int k(\cdot,x) f(x) \rho(\dv x).
	\end{gather}
	One can view the usual hypothesis class in kernel ML methods as 
	the image
	of the unit ball under this operator \citep{Williamson:2001aa}.
	But Markov kernels can also be written in a similar form.  As 
	\citet[pages 38 and 46]{Cinlar:2011aa} observes, we can express a
	Markov kernel $K\colon Y\expto X$ as 
	\begin{gather}
              K(y,\dv x)=\rho(\dv x) k(y,x),
	\end{gather}
	where $k$ is known as a kernel density relative to the reference
	measure $\rho$ and the operation of
	$K$ on a function $f\colon X\rightarrow\R$ can be written as 
	\begin{gather}
		\label{eq:markov-kernel-visa-density}
		Kf(\cdot)=\int K(\cdot,\dv x) f(x) = \int k(\cdot,x)
		f(x)\rho(\dv x).
	\end{gather}
	Comparing \eqref{eq:integral-operator} and
	\eqref{eq:markov-kernel-visa-density} we see that the Markov kernel
	performs a similar smoothing operation to $T_k$. When one
	takes account of Theorem \ref{prop:I-F-ET}, one concludes that 
	the choice of a kernel in a kernel learning machine is in effect an
	hypothesis about the type of noise the observations will be
	affected by. For example, using a Gaussian translation invariant
	kernel is an inductive bias which implicitly assumes the $\Xsf$
	measurements are corrupted by additive Gaussian noise.  (This last
	statement is perhaps misleading; we stress that it is
	$\ell\circ\scr H$ which is smoothed by the kernel
	$K$, not $\scr H$ itself. Understanding the effect of $K$
	directly on $\scr H$ seems challenging.)
\end{remark}

\section{Conclusion}
\label{sec:conclusion}
\vspace*{-2mm}
\begin{flushright}
      \footnotesize
      \textit{
        There is no in-formation, only trans-formation.}\\ 
        --- Bruno Latour\footnote{See \citep[page
	149]{latour2007reassembling} and \citep{Lovink1997Latour}.
      }
\end{flushright}

Motivated by  the epigram at the beginning of the paper, we  have used
Grothendieck's ``relative method'', whereby one understands an object, not
by studying the object itself, but by studying its morphisms.
We have seen that by construing information processing as
a transformation on the \emph{type} of information, rather than as a
manipulation of the \emph{amount} of some fixed type of information, one
obtains new insights into the nature of information. 

In doing so, we formulated a substantial generalisation of information,
which subsumes existing measures, including $\phi$-divergences and MMD.
The $D$ and $\scr F$ informations also induce corresponding notions of entropy (see
Appendix \ref{sec:entropies}). Their naturalness is manifest by the general
bridge to Bayes risks and constrained Bayes risks.  By working with
the variational form in which we define them, we can readily determine the
effect of noisy observations. We have shown that for both label noise and
attribute noise, the effect of the noisy observations can be captured by a
change of the measure of information used. This leads to information
processing \emph{equalities} instead of the traditional \emph{inequalities}
(which themselves are one of the basic results in information theory,
underpinning the notion of statistical sufficiency).  The new measures of
information provide insight into the variational representation of
$\phi$-divergences, as well as a new interpretation of the choice of kernel
in SVMs and MMD. 

The bridge results 
offer a way to avoid duplicate analytical work: 
for example, one does not need to separately
analyse the estimation properties of statistical divergences
\citep{Sreekumar2022}; one can simply convert to the equivalent statistical
decision problem for which many results already exist. In light of the
bridge result, one should hardly be surprised that  the constrained
variational representation of $\phi$-divergence has generalization
performance controlled by the  Rademacher complexity of the discriminator
set  \citep{Zhang2017}.

The information processing equality for $\scr F$-information generalises an
insight developed by \citet{Bishop:1995aa}  that the 
addition of noise (in training) is equivalent to a form of
regularization\footnote{Bishop's result is not quite the whole story, as
	explained by \citet{An:1996wt}. But the general conclusion is
	correct: adding noise to the input data encourages the learned model to
	be smoother than it would have been otherwise; confer 
	\citep{Grandvalet:1997aa}}. 
It goes beyond Bishop's result in that it applies to any ``noise'' (not 
necessarily additive) and explains the effect of
``adding'' noise precisely in terms of the 
effect on the hypothesis class.

The information processing results in the paper differ from the classical ones
in that they change the measure of information used. 
This is metaphorically changing the ``ruler'' used to
measure information on either side of the noisy channel.  The bridge
between information and expected loss shows that there is no reason to
expect there is a single canonical measure of information (as soon as one
accepts there is no single canonical loss function). 

Taken as a whole, the results show that at least for questions 
relating to prediction and learning, it makes no sense to talk of
``the'' information in one's data. While it is widely accepted that 
different problems demand different loss functions, it is also often
assumed that Shannon information is \emph{the} only measure of 
``information'' available\footnote{This is a point well 
	acknowledged by information theorists:
\begin{quote}
	[T]he fact that entropy has been proved in a meaningful sense to be
	the unique correct information measure for the purposes of
	communication does not prove that it is either unique or a correct
	measure to use in some other field in which no issue of encoding or
	other changes in representation arises \citep[page
	500]{Elias:1983aa}.
\end{quote}
}. For example \citet{rauh2017coarse} make much 
of the fact that although the worst Bayes risk (over all losses)
of an experiment may be made worse after passing through a channel,
particular measures of information may not be degraded at all.
Given the bridge between risks and measures of information, 
this can be seen as simply a mistake about quantification; 
the Blackwell-Sherman-Stein (BSS) theorem (recall Remark \ref{rem:bss}) 
to which they appeal, is stated 
in terms of either all loss functions or all measures of information.
Similarly, in much recent work in ML, the choice of a particular measure
of information is taken to be essentially one of convenience, and not related
to the underlying problem to be solved (in the way that one's choice of loss
function ideally is). The results of the paper show that choosing one's
measure of information is literally equivalent to choosing one's loss
function in a statistical decision problem, and thus is significant,
consequential, and not a mere matter of convenience or convention.


\subsection*{Acknowledgements}
RW's contribution was funded in part by the Deutsche Forschungsgemeinschaft
under Germany’s Excellence Strategy –-
EXC number 2064/1 –- Project number 390727645.  This work was presented (as
``Data Processing Equalities'') at the Tokyo workshop on Deep Learning:
Theory, Algorithms, and Applications (March 2018) \citep{tokyo:aa}, 
and at the Information
Theory in Machine Learning workshop, NeurIPS December 2019
\citep{ITML19:aa}.  A special
case of the argument in Example \ref{ex:attribute-noise} was developed by
RW in April 2007 [sic] after a discussion with Arthur Gretton.  An earlier
version of the proof of Proposition \ref{prop:limID}  
was developed by Etienne de Montbrun.
Thanks to Zak Mhammedi and Aneesh Barthakur for comments and corrections
and to Kamalaruban Parmeswaran and Brendan van Rooyen for discussions and
questions.

\appendix

\section{The \texorpdfstring{$\phi$}{phi}-divergence and its
         Variational Representation}
\label{sec:var-rep}

In this appendix we present some facts concerning the classical 
$\phi$-divergences and its variational representation and their relationship 
to our $D$- and $\mathcal{F}$-informations.

\subsection{Some examples of \texorpdfstring{$D_\phi$}{Dphi}}

\begin{figure}[t]
   \begin{center}
     \includegraphics[width=0.32\textwidth]{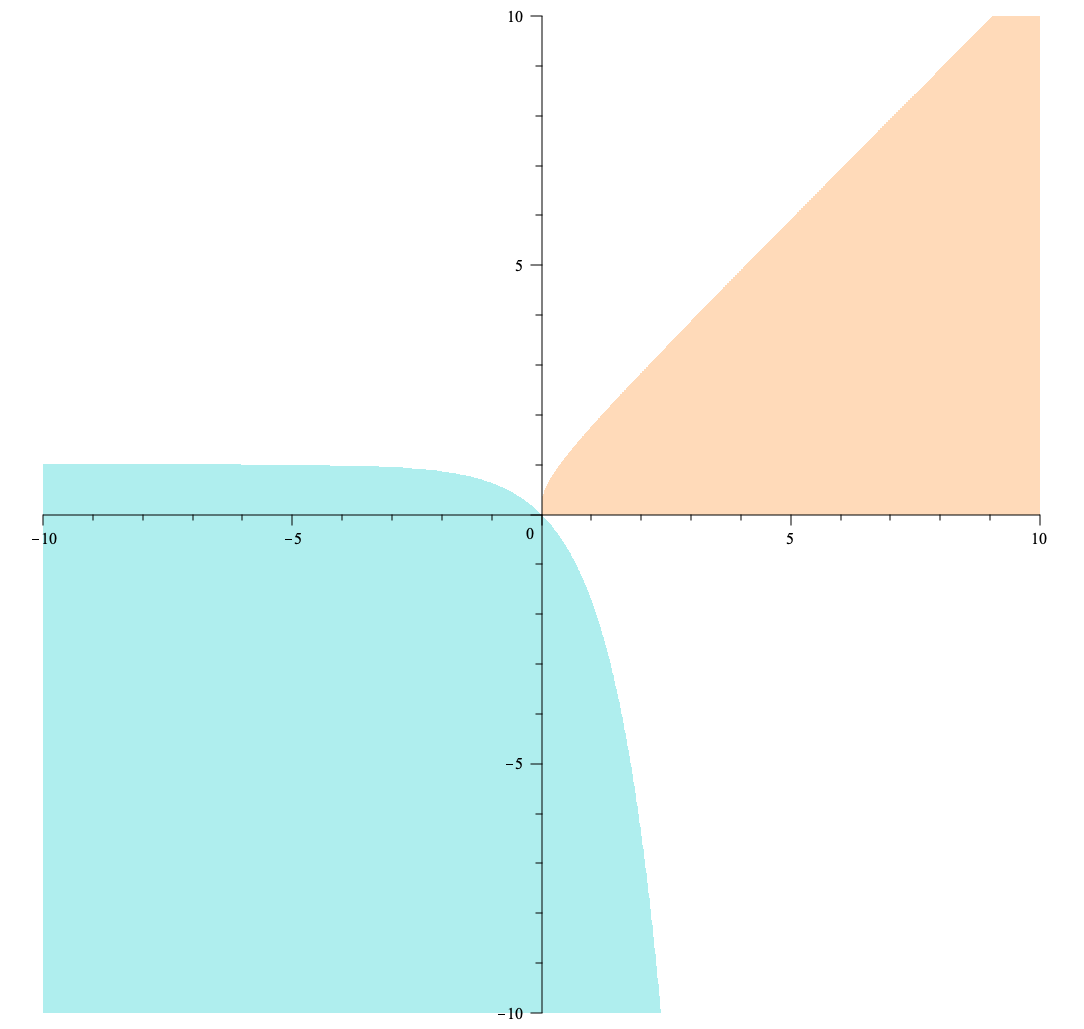}
     \includegraphics[width=0.32\textwidth]{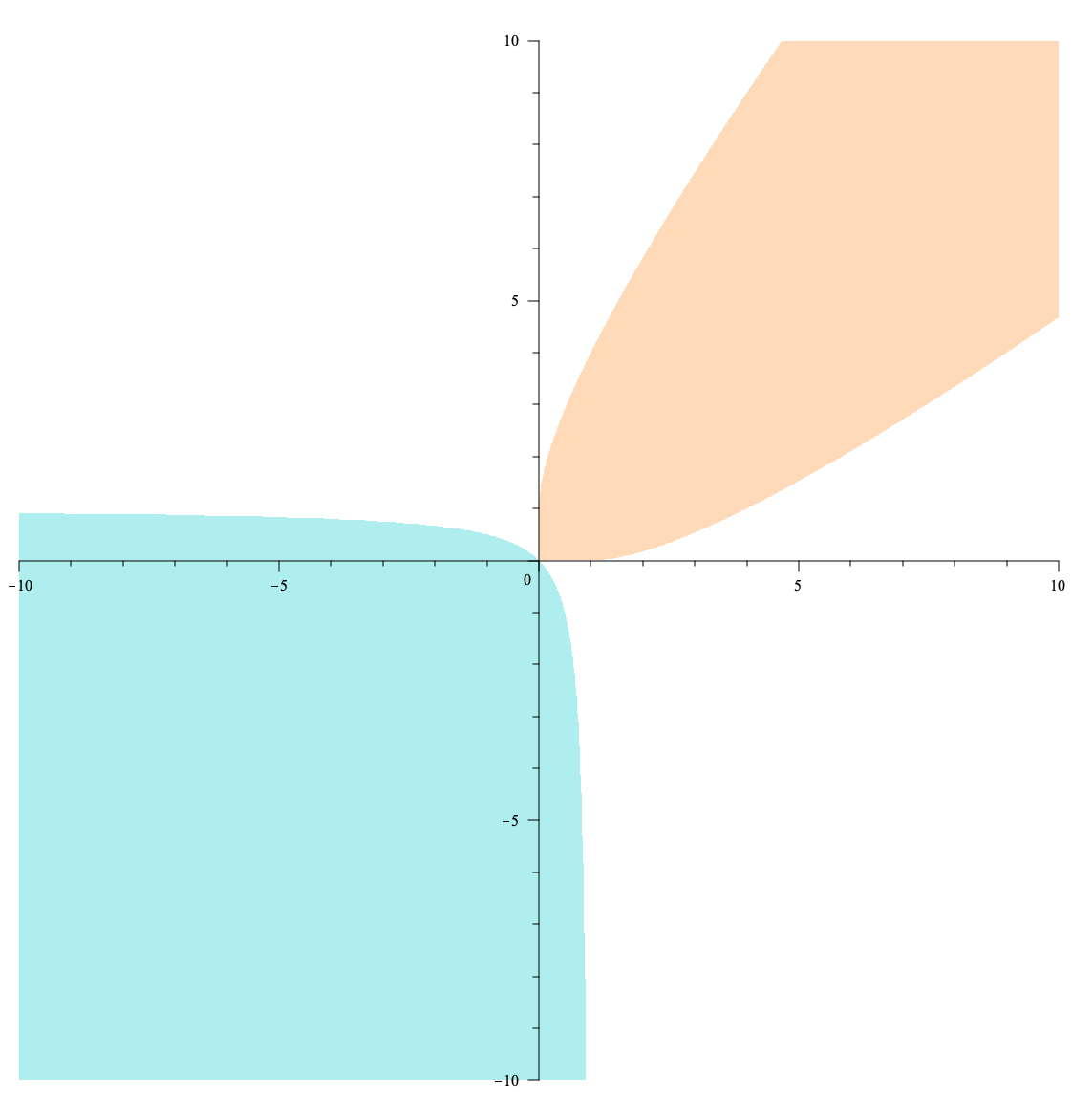}
     \includegraphics[width=0.32\textwidth]{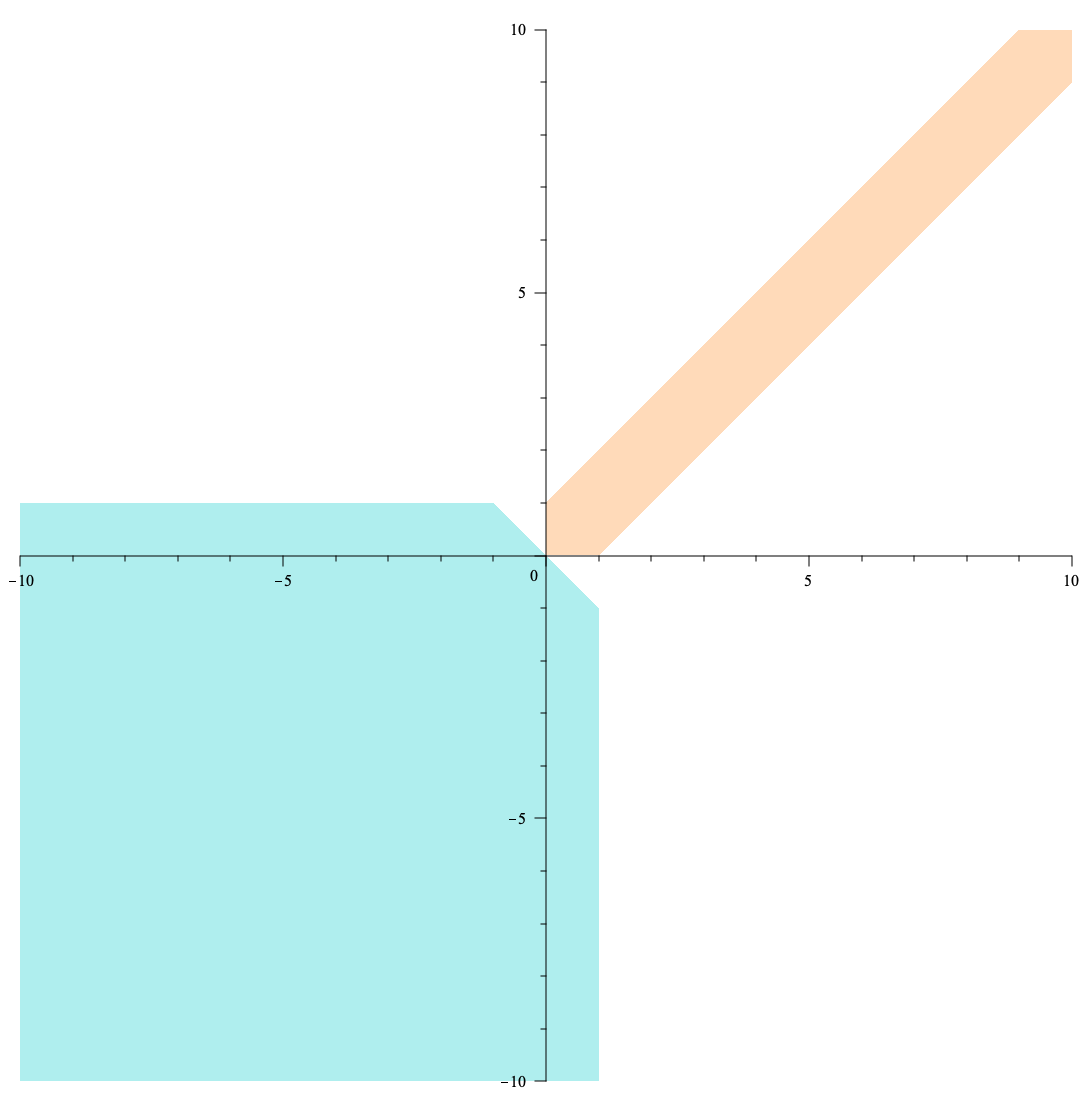}
   \end{center}
   \caption{$D_\phi$ (Turquoise)  and $C_\phi=(D_\phi)^\polar$ (Peach) for
     $\phi_{\mathrm{KL}}$, $\phi_{\mathrm{Hell}}$ and
     $\phi_{\mathrm{Var}}$ (left to right). Since in all cases
     $D_\phi$ and $C_\phi$ are unbounded sets, we have plotted
     their restriction
     to $[-10,10]^2$.  The purpose of including the polars $(D_\phi)^\polar$ 
     is motivated in Appendix \ref{sec:expected-gauge} \label{figure:D-C}.}
 \end{figure}

When $n=2$, we can compute some examples for classical $\phi$ divergences;
see Table \ref{table:phi-divergences}.
Figure \ref{figure:D-C} illustrates $D_\phi$ and $C_\phi=(D_\phi)^\polar$ for three
different $\phi$
(for such figures, it is helpful to use
$ (-\hyp \phi^*)^\polar = \lev_{\le 1} \breve{\phi}$).

\newcommand{\phant}{$\phantom{\displaystyle\int}$}
\begin{table}[t]
	\scriptsize
   \begin{center}
     \begin{tabular}{lll}
       \rowcolor[gray]{0.15}
       {\color{white} Divergence name}\phant               & {\color{white}
       $\phi(x)$, $x\ge 0$}                                &
       {\color{white} $\phi^*(x)$, $x\in\R$}                                                                 \\[1mm]
       \rowcolor[gray]{0.90}
       Variational \phant                                  & $\phi_{\mathrm{Var}}(x)=|x-1|$                &
       $-\iver{x<-1}+\iver{-1\le x\le 1}x
       +\iver{x>1}\infty$                                                                                    \\[1mm]
       \rowcolor[gray]{0.80}
       Kullback-Leibler \phant                             &
       $\phi_{\mathrm{KL}}(x)=x\log(x) -x +1$              &
       $e^x-1$                                                                                               \\[1mm]
       \rowcolor[gray]{0.90}
       Squared Hellinger \phant                            &
       $\phi_{\mathrm{Hell}}(x)=\rbr{\sqrt{x}-1}^2$ &
       $\iver{x\le 1} \frac{x}{1-x}+
       \iver{x>1}\infty$                                                                                     \\[1mm]
       \rowcolor[gray]{0.80}
       Chi-squared\phant                                   & $\phi_{\chi^2}(x)= (x-1)^2$                   &
       $\frac{1}{4} x^2 +x$                                                                                  \\[1mm]
       \rowcolor[gray]{0.90}
       Jensen-Shannon\phant                                &
       $
         \phi_{\mathrm{JS}}(x)=x\log x -(x+1)\log\frac{x+1}{2}$
       \ \                                                 & $-\iver{x\le\log 2 }\log(2-e^x)
       +\iver{x>\log 2 }\infty$                                                                              \\[1mm]
       \rowcolor[gray]{0.80}
       Triangular                                          & $\phi_{\mathrm{Tri}}(x)=\frac{(x-1)^3}{x+1} $
                                                           & $-\iver{x\le -3} +\iver{1< x}\infty$            \\
       \rowcolor[gray]{0.80}
                                                           &                                               &
       \hspace*{5mm}
       $+\iver{-3\le x\le
           1}(\sqrt{1-x}-1)(\sqrt{1-x}-3)$
     \end{tabular}
   \end{center}
   \caption{Common $\phi$-divergences and their associated $\phi^*$;
     drawn from \citep{Terjek:2021aa} which has further examples.
     In all cases, $\phi^*(0)=0$, and thus
     $\hyp(-\phi^*)\in\Dfrak^2$. We adopt the
     convention that $\iver{\mbox{false}}\cdot \infty=0$.
     \label{table:phi-divergences} }
 \end{table}

\subsection{The variational representation of 
	\texorpdfstring{$\phi$}{phi}-divergences}
 \label{sec:variational-representation}
 When $n=2$ ($Y=[2]$) we can relate $\I_{\scr{F}}$ to the 
 Csisz\'{a}r divergence $\I_\phi$.
For $\phi\in\Phi$ let
\begin{gather}
\Def{\scr{F}_\phi}{\measf(X,\hyp(-\phi^*))}.
\end{gather}
Proposition \ref{thm:fdiv_equivalence} and Theorem \ref{thm:fscr_brisk_bridge}
immediately imply
\begin{gather}
  \label{eq:three-informations}
  {
    \color{ForestGreen}
    \I_\phi(E)  = \I_{\scr{F}_\phi}(E).
  }
\end{gather}
It is now instructive to
relate $\I_{\scr{F}_\phi}(E)$ to the variational representation of
$\I_\phi$ \citep{Keziou2003}\footnote{
	Such representations have attracted some attention recently; for
	example
	\citep[section 4.3]{Agrawal2021} and 
	\citep{Terjek:2021aa, Ruderman:2012tq, Birrell:2022vm}.  A focus of
	these works is to develop restrictions on the class of functions one
	optimises over in order to aid their statistical estimation; the
	$\mathcal{F}$-information of the present paper can be seen to embrace a
	similar philosophy.},
which is central to the concept of $f$-GANS
\citep{Nowozin:2016aa}.

\begin{proposition}\label{prop:iscript_conjugate_result}
  Suppose $\phi\in\Phi$, $Y=[2]$, and $E: Y\expto X$.
  Then
  \begin{gather}
    \label{eq:binary-variational-representation}
    {
      \color{ForestGreen}
      \I_{\scr{F}_\phi}(E)=
      \sup_{g\in\measf(X,\R)} \left(\E_{E_1} g -
      \E_{E_2} \phi^*\circ g\right).
    }
  \end{gather}
\end{proposition}
\begin{proof}
  Using \eqref{eq:I-F-via-rho} we have
  \begin{align*}
    \I_{\scr{F}_\phi}(E) & =\sup_{f=(f_1,f_2)\in\scr{F}_\phi}
    \E_\rho\left\langle \left(\frac{\dv E_1}{\dv\rho},\frac{\dv
    E_2}{\dv\rho}\right),(f_1,f_2)\right\rangle                              \\
                         & = \sup_{\{(f_1,f_2)\st f_2\le -\phi^*\circ f_1\}}
    \E_\rho\left\langle \left(\frac{\dv E_1}{\dv\rho},\frac{\dv
    E_2}{\dv\rho}\right),(f_1,f_2)\right\rangle                              \\
                         & =\sup_{f_1\in\measf(X,\R)}
    \E_\rho\left\langle \left(\frac{\dv E_1}{\dv\rho},\frac{\dv
    E_2}{\dv\rho}\right),(f_1,-\phi^*\circ f_1)\right\rangle,                \\
    \intertext{since $\frac{\dv E_2}{\dv\rho}>0$ and we can restrict the
      optimization to be such that
      $f_2(x)=-\phi^*(f_1(x))$ for all
      $x\in X$, and we exploited the fact that $\scr{F}_\phi$ is the set of
      \emph{all} measurable functions mapping into $\hyp(-\phi^*)$,}
                         & = \sup_{f_1\in\measf(X,\R)}
    \E_\rho\left(\frac{\dv E_1}{\dv\rho} f_1 - \frac{\dv E_2}{\dv\rho}
    \phi^*\circ f_1\right)                                                   \\
                         & = \sup_{f_1\in\measf(X,\R)} \int_X \left(f_1
    \frac{\dv E_1}{\dv\rho} - \phi^*\circ f_1 \frac{\dv E_2}{\dv\rho}
    \right)\dv\rho                                                           \\
                         & = \sup_{f_1\in\measf(X,\R)}\left( \int_X
    f_1\frac{\dv E_1}{\dv\rho}\dv\rho - \int_X
    \phi^*(f_1)\frac{\dv E_2}{\dv\rho}\dv\rho\right)                         \\
                         & =  \sup_{f_1\in\measf(X,\R)}\left(\int_X
    f_1\dv E_1 - \int_X \phi^*(f_1)\dv E_2\right)                            \\
                         & = \sup_{f_1\in\measf(X,\R)}
    \rbr{\E_{E_1} f_1 - \E_{E_2} \phi^*\circ f_1}.\qedhere
  \end{align*}
\end{proof}

It is apparent from the proof of the above that the asymmetry in the usual
variational representation
\eqref{eq:binary-variational-representation}, whereby $\phi^*$ appears
in only one of the terms, arises from the choice of
$E_2$ as the dominating measure and the parametrisation of
$D\in\Dfrak^2$ by
$\phi\in\Phi$. Such a choice is problematic if $E_2$ does
not dominate $E_1$, leading to less elegant general definitions being
necessary for $\I_\phi$ \citep{Liese2008,LieVaj06}. The one
advantage of \eqref{eq:binary-variational-representation} over
\eqref{defn:d_information} when $Y=[2]$ is that the optimisation is over
$\R$-valued functions rather than $\R^2$-valued functions. However,
as seen in Section \ref{sec:information-defs}, the symmetric representation
\eqref{defn:d_information} has significant advantages in
understanding
the effect of the product of experiments (in the form of observation
channels).

When $\phi=\Def{\phi_{\mathrm{var}}}{t\mapsto|t-1|}$,
$\I_{\phi_{\mathrm{var}}}$ is known as the \Def{variational divergence}
which is examined in detail in \S\ref{sec:variational-divergence}. Finally, the form of \eqref{eq:binary-variational-representation} suggests the variant
\[
  \I_{\scr H}(E)=
  \sup_{h\in\scr H} \rbr{\E_{E_1} h - \E_{E_2} \phi^*\circ h},
\]
where $\scr H\subsetneq\measf(X,\R)$. The
functional $\I_\scr H$ is what is estimated in practice by virtue of choice
of a suitable class over which to empirically optimise
\eqref{eq:binary-variational-representation}, often replacing
${E_i}$ by their empirical approximations ${\hat{E}_i^m}$,
where for $A\in\Sigma_X$, $\Def{\hat{E}_i^m(A)}{\frac{1}{m}\sum_{j\in[m]}\iver{A\ni
      x_j}}$.

An alternate way of expressing the general form of $\I_{\scr F}(E)$ that
is similar to the classical variational representation of a binary
$\phi$-divergence is given below.
Let $\scr H\subseteq \measf(X,\probm([n]))$. For $D\in\Dfrak^n$, assume
there is a measurable selection
$\nabla\sprt_D\in\subdiff\sprt_D$ (confer Proposition
\ref{prop:dinf_witness}). We can thus write
\begin{gather} \label{eq:I-gain-D-H}
  \I_{\nabla\sprt_D\circ\scr H}(E) = \sup_{h\in \scr H} \E_\rho \
  \left\langle\frac{\dv E}{\dv \rho}, \nabla\sprt_D\circ
  h\right\rangle.
\end{gather}
Observe that $\nabla\sprt_D\circ\scr H=\{\nabla\sprt_D\circ h\st
  h\in\scr H\}\subseteq\measf(X,D)$.
This is a way to use classes of functions mapping to $\R^n$ in an
elegant manner to define a restricted version of $\I_D$.
Observe that \eqref{eq:I-gain-D-H} is symmetric in the appearance of
$\nabla\sprt_D$, in a manner that \eqref{eq:binary-variational-representation} is
not, but one needs to work with vector valued functions
$h: X\to\R^n$. Given a function class
${\cal R}\subseteq\measf(X,\R)$,  one could  induce
$\Def{\scr H_{\cal R}}{\{X\ni x\mapsto (r_1(x),\ldots,r_n(x)) \st r_i\in{\cal R}\ \forall
    i\in[n]\}}$, allowing us to define
$\Def{\I_{{\cal R},D}(E)}{\I_{\nabla\sprt_D\circ\scr H_{\cal R}}(E)}$.

\subsection{The Variational Divergence}
\label{sec:variational-divergence}
The binary Variational divergence has
$\phivar(u)=|u-1|$ for $u\in\R+$.
\begin{lemma}
  The Legendre-Fenchel conjugate of $\phivar$ is given by
  \[
    \phivar^*(s)=\begin{cases} s & s\in[-1,+1] \\ +\infty & s\not\in
              [-1,+1].\end{cases}
  \]
\end{lemma}
\begin{proof}
  We have
\vspace*{-3mm}
  \begin{align*}
    \phivar^*(s) & = \sup_{u\ge 0} u\cdot s-\phivar(u)                                                \\
              & = \sup_{u\ge 0} u\cdot s - \begin{cases} u-1 & u\ge 1\\ 1-u & u\le 1\end{cases} \\
              & =\sup_{u\ge 0} \begin{cases} u(s-1)+1 & u\ge 1 \\ u(s+1)-1 & u\le
              1.\end{cases}
    \intertext{Suppose $s>1$. Then the supremum is attained for $u\ge 1$ and
      $\phivar^*(s)=\sup_{u\ge 0} uc+1$ for $c>0$
      which equals $+\infty$. Similarly if $s<-1$, the supremum is attained for
      $u\le 1$ and again $\phivar^*(s)=+\infty$. Suppose $s\in[-1,1]$; we have}
    \phivar^*(s) & = \max\left\{ \sup_{u\ge 1} u(s-1)+1,\ \sup_{u\in[0,1]}
    u(s+1)-1\right\}                                                                            \\
              & = \max\{ 1(s-1)+1,\ 1(s+1)-1\}                                                  \\
              & = s\ \ \ \ \ \ \ \ \ \ \ s\in[-1,1],
  \end{align*}
  which completes the proof.
\end{proof}
Now consider  the evaluation of
\begin{align}
  \I_{{\scr F}_{\phivar}}(E) & = \sup_{g:  X\to\R}
  \E_{E_1} g - \E_{E_2} \phivar^*\circ g.\nonumber                         \\
  \intertext{If, for any $x\in X$, $g(x)\not\in [-1,+1]$, then the second term will be
    infinite which will push the whole value to $-\infty$. Thus the sup can
    never be attained if $g$ takes on values outside of $[-1,+1]$ (except on a
    $E_2$-negligible set). Hence we need only consider}
  \I_{{\scr F}_{\phivar}}(E) & = \sup_{g:  X\to
  [-1,1]} \E_{E_1} g - \E_{E_2}  g\nonumber                             \\
  \intertext{Since the objective is linear, and the constraint set convex, the supremum
    is attained at the boundary and hence}
  \I_{{\scr F}_{\phivar}}(E) & = \sup_{g:  X\to
    \{-1,1\}} \E_{E_1} g - \E_{E_2} \phivar^*\circ
  g\nonumber                                                            \\
                          & = 2 \sup_{g:  X\to
    \{0,1\}} \E_{E_1} g - \E_{E_2} \phivar^*\circ g
  \label{eq:variational-intermediate}                                   \\
                          & = 2 \sup_{A\in\Sigma_X} E_1 (A) -
  E_2(A)\label{eq:classical-variational-form-no-abs}                    \\
                          & = 2 \sup_{A\in\Sigma_X} |E_1 (A) - E_2(A)|,
  \label{eq:classical-variational-form}
\end{align}
where the last step is shown in \citep{Strasser:1985aa}.
Observe that \eqref{eq:variational-intermediate} can also be written as
\begin{gather}
  2 \sup_{g:  X\to
    [0,1]} \E_{E_1} g - \E_{E_2} \phivar^*\circ g
\end{gather}

We now determine
$\Def{\Dvar}{\hyp(-\phivar^*)}$.
\begin{lemma}
  \label{lemma:Dvar-explicit}
  Let $\Def{H_{n,r}^-}{\{x\in\R^n\st \inp{ n,x} -r \le
      0\}}$ denote the negative halfspace with normal vector $n$ and offset $r$.
  The set $\Dvar$ can be written
  \begin{gather}
    \Dvar= H_{1_2,0}^- \cap H_{\e_1,1}^- \cap H_{\e_2,1}^- .
    \label{eq:Dvar-explicit}
  \end{gather}
\end{lemma}
\begin{proof}
  We have
  \begin{align}
    \hyp(-\phivar^*) & = \setcond{(x,y)\in\R^2}{ y\le -\phivar^*(x)}\nonumber                            \\
                  & = \left\{(x,y)\in\R^2\st y\le\begin{cases} -x & x\in[-1,+1]      \\ -\infty
                 & x\not\in [-1,+1]\end{cases}\right\}\nonumber \\
                  & =\setcond{(x,y)\in\R^2}{ x\in[-1,+1], y\le -x}.\nonumber
  \end{align}
  Now Lemma
  \ref{lem:dinf_conv_closure} implies
  $\I_D(E)=\I_{\co D}(E)$, and hence we can take the
  convex hull of the above to obtain
  \begin{align}
    \Dvar & = \setcond{(x,y)\in \R^2}{ y\le 1} \cap \{(x,y)\in\R^2\st y\le
    -x\} \cap \setcond{(x,y)\in\R^2}{ x\le 1}\nonumber                           \\
          & = \{(x,y)\in\R^2\st y\le 1 \mbox{\ and\ } x\le 1 \mbox{\ and\ } y\le
    -x\}.\nonumber
  \end{align}
  We can thus more compactly write $\Dvar$ as in
  \eqref{eq:Dvar-explicit}.
\end{proof}

Lemma \ref{lemma:Dvar-explicit} suggests the following generalisation which we now take as a
definition
\begin{gather}
  \Def{\Dvarn}  \defeq H_{1_n,0}^- \cap\bigcap_{i\in[n]} H_{\e_i,1}^- .
  \label{eq:Dvarn}
\end{gather}
Observe that $\Dvarn$ is the maximal (by set inclusion) element of
$\cvxrec(\R^n, \R-^n)$ satisfying the normalisation condition
$\bigvee_{i\in[n]}\sprt_D(\e_i)=1$.
We now compute $\sprt_{\Dvarn}$.
\begin{lemma}
  The support function of $\Dvarn$ is given by
  \[
    \sprt_{\Dvarn} (x) =\sum_{i\in[n]} x_i -n \bigwedge_{j\in[n]} x_j .
  \]
\end{lemma}
\begin{proof}
  Note that $\Dvarn$
  is a intersection of half spaces and thus its support function is the same
  as the support function of its extreme points, which is the union of the
  $n$ vertices created.  Denote
  the vertices $v_j$ for $j\in[n]$. We have
  \begin{align}
    v_j & = H_{1_n,0}^- \cap  \bigcap_{i\in[n]\setminus\{j\}}
    H_{\e_i,1}^- \nonumber                                          \\
        & = \{x\in\R^n\st \inp{ x ,\,\e_i} -1 =0\ \forall i\ne
    j\mbox{\ and\ } \inp{ x,\,1_n}=0\}\nonumber                     \\
        & =  \left\{x\in\R^n\st x_i=i\ \forall i\ne j\mbox{\ and\ }
    \textstyle\sum_{k\in[n]} x_k =0\right\}\nonumber                \\
        & =  \{x\in\R^n\st x_i=1\ \forall i\ne j\mbox{\  and\ }
    x_j=-(n-1)\}\nonumber                                           \\
        & = 1_n- n \e_j. \label{eq:vj}
  \end{align}
  Thus  for $x\in\R+^n$, using 
  \citep[Theorem C.3.3.2 (ii)]{hiriarturruty2001fca} we have 
  $\sprt_{\clco(\cup_{j\in[n]}
      \{v_j\}}=\sup_{j\in[n]}\sprt_{\{v_j\})}$ and hence
  \begin{align*}
    \sprt_\Dvarn(x) & = \sup_{j\in[n]} \langle 1_n - n\e_j,\,
    x\rangle                                                           \\
                    & = \inp{ 1_n,\, x} +n \sup_{j\in[n]} - x_j        \\
                    & = \sum_{i\in[n]} x_i -n \bigwedge_{j\in[n]}
		    x_j.\qedhere
  \end{align*}
\end{proof}

We can now determine an explicit expression for $\I_\Dvarn(E)$.
Let $(\bar{X}_1,\ldots,\bar{X}_n)$ be a measurable partition of $X$
(i.e.~$\bar{X}_i$ are measurable for $i\in[n]$)  defined via
\begin{gather} \label{eq:Xbar-partition}
  \Def{\bar{X}_i}{\left\{x\in X \st \textstyle
    \min\left(\operatornamewithlimits{Argmin}_{j\in[n]}
    \frac{\dv\, E_j}{\dv\,\rho}(x) =i\right)\right\}}.
\end{gather}
(The additional $\min$ is to break ties.)
It is immediate that this is indeed a partition of $X$, i.e.
$\bigcup_{k\in[n]} \bar{X}_k = X$ and $\bar{X}_i
  \cap\bar{X}_j=\varnothing$ for $i\ne j$.
Consequently
\begin{align}
  \I_\Dvarn(E) & = \int_X \sum_{i\in[n]} \frac{\dv E_i}{\dv\rho}(x)
  \rho(\dv x) - n\int_X \bigwedge_{j\in[n]} \frac{\dv E_j}{\dv \rho}(x)
  \rho(\dv x)\nonumber                                                          \\
               & =  \sum_{i\in[n]}\int_X \dv E_i - n\int_X
  \bigwedge_{j\in[n]} \frac{\dv E_j}{\dv\rho}(x) \rho(\dv
  x)\label{eq:partition-integral}                                               \\
               & = n\left[ 1 - \sum_{k\in[n]}\int_{\bar{X}_k} \frac{\dv
  E_k}{\dv\rho}(x) \rho(\dv x)\right]\nonumber                                  \\
               & =  n\left[ 1 - \sum_{k\in[n]} E_k (\bar{X}_k)\right],\nonumber \\
  \shortintertext{using the properties of the partition
    $(\bar{X}_1,\ldots,\bar{X}_n)$.}
\end{align}
Observe that choosing any other partition of $X$ would result in a larger
value of the second integral in \eqref{eq:partition-integral} and thus a smaller
value for the overall expression. Thus if ${\cal P}_n(X)$  denotes the set of
all measurable $n$-partitions of $X$, we can write
\begin{gather}
  \I_\Dvarn(E) = \sup_{(X_1,\ldots,X_m)\in{\cal P}_n(X)} n\left[ 1-
    \sum_{k\in[n]} E_k(X_k)\right].
\end{gather}
When $Y=[2]$, we obtain
\begin{align*}
  \I_{D_{\mathrm{var}}^{(2)}}(E) & =
  \sup_{(X_1,X_2)\in{\cal P}_2(X)} 2 [1-(E_1(X_1)+E_2(X_2))]          \\
                                 & = 2 \sup_{\substack{X_1\subseteq X \\ X_1\ \mathrm{measurable}}}
  [1-(E_1(X_1)+E_2(X\setminus X_1))]                                  \\
                                 & = 2 \sup_{\substack{X_1\subseteq X \\ X_1\ \mathrm{measurable}}} [1-(E_1(X_1)+1- E_2(X_1))]\\
                                 & = 2 \sup_{\substack{X_1\subseteq X \\ X_1\ \mathrm{measurable}}}  [E_1(X_1)- E_2(X_1)],
\end{align*}
which can be recognised as being equivalent to
\eqref{eq:classical-variational-form-no-abs}.

Finally we observe a special case of \eqref{eq:IDSE} for $D=\Dvarn$ when
$S$ takes the particular symmetric form $S_\alpha$ where the $j$th
column of $S_\alpha^*$ is $s_j^*=\alpha \e_j +\frac{1-\alpha}{n}1_n$.
When $\alpha=1$ this is the identity matrix, and for $\alpha\in[0,1]$ it
corresponds to the observation channel providing the correct label with
probability $\alpha$ and with probability $1-\alpha$
a label chosen at random from $[n]$ is chosen (which could in fact be correct). The
set $S_\alpha^* \Dvarn$ can be readily determined by exploiting the fact we
need only determine its support function $\sprt_{S_\alpha^* \Dvarn}(x)$
for $x\in\R+^n$. Thus we can exploit \eqref{eq:vj} and we need
only compute (for $j\in[n]$)
\begin{align*}
  S_\alpha^* v_j & = 1_n - n\left(\alpha \e_j +
  \frac{1-\alpha}{n}1_n\right)                  \\
                 & =\alpha(1_n - n\e_j)         \\
                 & = \alpha v_j.
\end{align*}
Thus $\sprt_{S_\alpha^* \Dvarn} = \alpha\sprt_D$ and so for any
$\alpha\in[0,1]$ and any $n$, we have the homogeneous relationship
\[
  {
      \color{ForestGreen}
      \I_\Dvarn(S_\alpha E) = \alpha \I_\Dvarn(E),
    }
\]
which we note has the \emph{same} measure of information on either side of the
equality (analogous to the typical strong data processing \emph{in}equalities one
finds in the literature).

\section{\texorpdfstring{$D$}{D}-Information as an Expected Gauge Function}
\label{sec:expected-gauge}
Classical binary information ``divergences'' are sometimes supposed to be
``like'' a distance (a metric). In this appendix we show that there is an 
element of truth in this supposition.  Metrics (as a formal notion of
``distance'')  are often (not always) induced by norms, and 
norms are particular examples of convex gauge functions 
(Minkowski functionals). 
In this appendix we show that it follows almost immediately from 
our definition of $D$-information that it is indeed an 
\emph{expected} gauge function, albeit one where the associated 
``unit ball'' of the gauge is neither symmetric nor compact.
The restriction of  $D\in\Dfrak$ allows an
insightful representation of $\I_D$ making use of the 
classical polar duality of closed convex sets containing the origin.

The \Def{conic hull} of a set $C\subset\R^n$ is
$\Def{\cone C}{(0,\infty)\cdot C}$. Given $C\in\cvx(\R^n)$, the \Def{polar} of $C$ is defined by
\[
	\Def{C^\polar}{\{x^*\st\forall x\in C,\ \langle x,x^*\rangle \le
	1\}}.
\]
We will make use of  the
following from \citep[Theorem 14.6]{Rockafellar:1970}:
\begin{proposition}
	\label{prop:polar-rec-cone}
	Suppose $C,C^\polar\in\cvx(\R^n)$ are a polar pair both containing
	the origin.  Then $(\rec C)^\polar = \cl\cone C^\polar$.
\end{proposition}

Given $C\in\cvx(\R^n)$, the \Def{gauge} of $C$ is defined by
\[
	\Def{\gamma_C(x)}{\inf\{\mu\ge 0 \st x\in \mu C\}}.
\]
Obviously given the gauge $\gamma_C$ one can recover $C$ via
$
	C=\lev_{\le 1} \gamma_C=\{x \st\gamma_C(x)\le 1\}.
$
(If $C$ is symmetric about the origin, then $\gamma_C$ is a norm.)
Let 
\[
	\Def{\Cfrak_0^n}{\left\{C\in\cvx(\R^n)\st\cone C\subseteq \R_{\ge 0}^n, \ 
	0\in \tbd C, \ \cone 1_n\subseteq C\right\}}.
\]
\begin{lemma}\label{lem:Dfrak-Cfrak}
	If $D\in\Dfrak_0^n$ then $D^\polar\in\Cfrak_0^n$.
\end{lemma}
\begin{proof}
	If $D$ is convex then so is $D^\polar$.
	By Proposition \ref{prop:polar-rec-cone}, since $0\in D$, $\rec D$ is the
	largest cone contained in $D$ and $(\rec D)^\polar = \cl\cone C$ is the
	smallest cone containing $C$. Thus when $\rec D=\{x\in\R^n \st \langle
	x,1_n\rangle \le 1\}$, $\cl\cone C=\{\alpha 1_n \st\alpha\ge 0\}$.
	Regardless of the choice of $D$, we always have $0\in \tbd D^\polar$.
	The final condition in the definition of $\Cfrak_0^n$ follows 
	since $D\subseteq C \Rightarrow D^\polar \supseteq C^\polar$, and  
	$(\lev_{\le 0} \langle,\cdot, 1_n\rangle)^\polar= \cone 1_n$.
\end{proof}

Gauges and support functions are dual to each other in the polar sense
	 \citep[Corollary C.3.2.5]{hiriarturruty2001fca}:
\begin{lemma}
	\label{lemma:gauge-support-duality}
	 Suppose $C\in\cvx(\R^n)$, then $\sigma_C=\gamma_{C^\polar}$.
\end{lemma}
\begin{proposition}
	For any $D\in\Dfrak^n$, $D^\polar\in\Cfrak^n$ and
	for any $E\colon [n]\expto X$,  and any reference measure $\rho$,
	\begin{gather}
		{\color{ForestGreen}
		\I_D(E) = \int_X \gamma_{D^\polar}\left(\textstyle\frac{\dv
		E}{\dv\rho}\right)\dv\rho.
	} \label{eq:I-D-as-gauge}
	\end{gather}
\end{proposition}
\begin{proof}
	The first claim  is just lemma \ref{lem:Dfrak-Cfrak}. 
	The second claim follows by applying
	Lemma \ref{lemma:gauge-support-duality} pointwise.
\end{proof}

\begin{figure}[t]
	\begin{center}
		\includegraphics[width=0.32\textwidth]{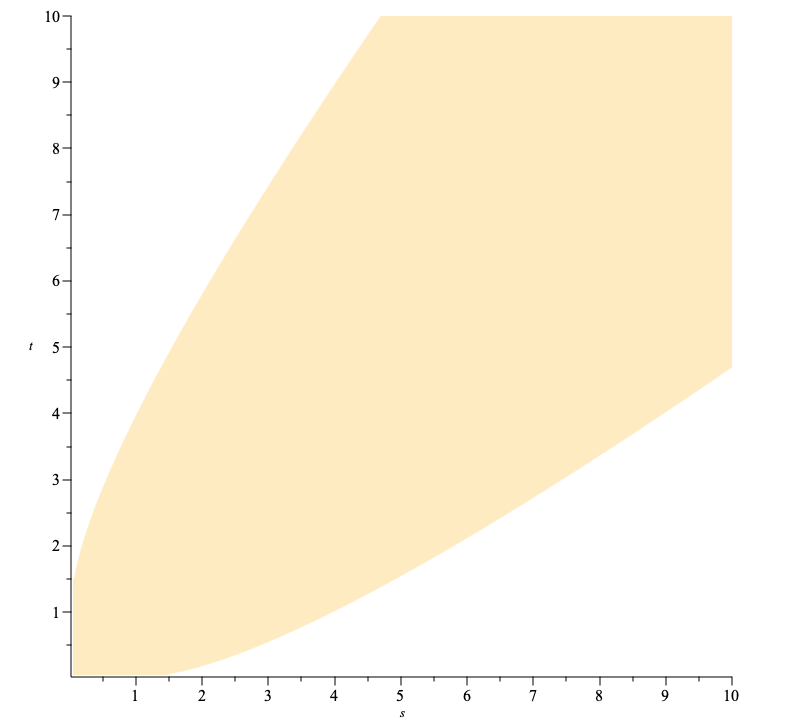}
		\includegraphics[width=0.32\textwidth]{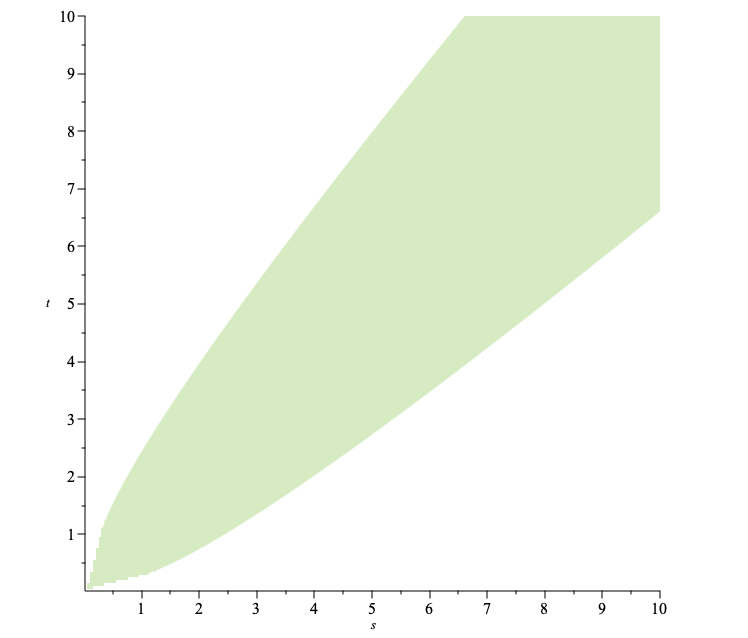}
		\includegraphics[width=0.32\textwidth]{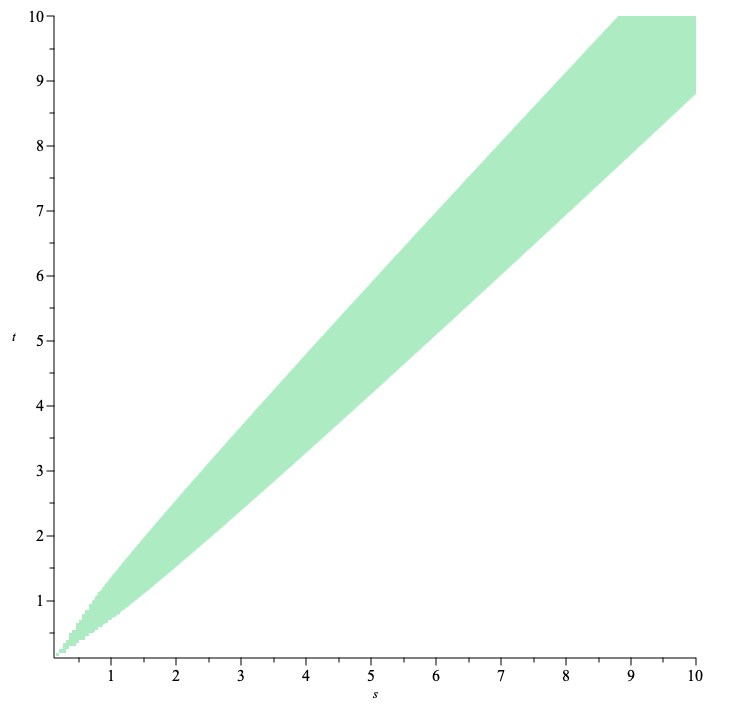}
	\end{center}
	\caption{The  
		corresponding polars for $R_r D_{\mathrm{Hell}}$ for
		$r=1,0.8, 0.6$ (restricted to
		$[0,10]^2$) corresponding to the  set-up as in Figure \ref{fig:C-D-r}. 
		Observe that for any $E\colon [2]\expto X$, as
		$r\downarrow 0.5$, the composition $R_r E$ approaches the totally 
		non-informative experiment $E^{\mathrm{tni}}$, and
		$\I_{R_r^*D}$ approaches what we might (oxymoronically) call the 
		totally noninformative information measure
		$\I^{\mathrm{tni}}=\I_{D_{\mathrm{tni}}}$, where
		$D_{\mathrm{tni}}=\{x\in\R^n \st \langle x,1_n\rangle\le 0\}$ and 
		$C_{\mathrm{tni}}=D_{\mathrm{tni}}^\polar=\{\alpha 1_n \st
		\alpha\ge 0\}$. The name is justified since
		$\I^{\mathrm{tni}}(E)=0$
		for all experiments $E$. \label{fig:polars}}
\end{figure}

Expressing $\I_D$ as an average of a gauge function as in
\eqref{eq:I-D-as-gauge} justifies the oft made
claim that divergence are ``like'' distances in some sense; the fact that
$D^\polar$ is not symmetric is why it is merely ``like''.  One can see that
$\I_D$ is ``gauging'' the average degree to which the vector $\frac{\dv
E}{\dv \rho}(x)$ is ``close'' to one of the canonical basis vectors $e_i$,
$i\in [n] $ since for $D^\polar\in\Cfrak_0^n$,
$\gamma_{D^\polar}(e_i)>0$ in that case. Conversely, since $\cl\cone
D^\polar=\{\alpha 1_n \st\alpha\ge 0\}$, we always have
$\gamma_{D^\polar}(1_n)=\sigma_D(1_n)=0$, corresponding to situations where
$\frac{\dv E}{\dv\rho}(x)=1_n$, and consequently it being impossible to
distinguish between the outcomes of the experiment at that $x$ --- in other
words a complete absence of ``information.''  Some example of polars of $D$ 
illustrated in Figure \ref{fig:polars}.

\section{Unconstrained and Constrained Entropies}
\label{sec:entropies}

Historically, the notion of the \emph{entropy} of a single distribution (or
random variable) preceded measures of information between two or more
distributions (or random variables)\footnote{%
    In classical thermodynamics, entropy has been taken to be the fundamental
    notion, with relative entropy (i.e. KL-divergence) as subsidiary. However
    recent work has shown that one can develop classical thermodynamics
    starting from relative entropy, with a number of advantages
    \citep{Floerchinger:2020uc}. They conclude  by speculating that it
    could be beneficial, for the foundations of thermodynamics, ``to think more
    often in terms of \emph{distinguishability} instead of \emph{missing
    information}'' \citep[page 11]{Floerchinger:2020uc};  
    confer \citep{ben2008farewell} which argued that ``missing information'' 
    was a better viewpoint than the classical ``degree of uncertainty'' 
    usually invoked to explain the intuition of physical entropy.}.
There is a large literature on different notions of entropy, starting 
with \citep{Shannon1948}, with $\phi$-entropies (analogous to 
$\phi$-divergences) specifically considered in
\citep{Csiszar1972,Daroczy:1970aa,Ben-Bassat:1978aa}.
In this appendix we recall how the entropy  of  a distribution
$\mu\in\probm(X)$ can be defined on the basis of comparison against a
``uniform'' measure $\upsilon\in\probm(X)$ cf.~\citep{Torgersen:1981,
Naudts:2008aa}.  Traditionally this comparison measure is taken for granted
as being Lebesgue measure, but we shall see it is an arbitrary choice and
the choice matters\footnote{This idea that unary properties are
intrinsically relative to some implicit reference has been developed for
the notion of Lorenz curves \citep{Buscemi:2017wn}, themselves related to ROC
curves
\citep{Schechtman:2019wu} which are intimately
related to certain families of $\I_D$ \citep[\S6.1]{Reid2011}.}.

Given $\mu\in\probm(X)$, define the experiment $E_\mu^\upsilon: [2]\expto X$ via
\begin{gather}
	\Def{E_\mu^\upsilon(1,\cdot)}{\mu(\cdot)} \mbox{\ \ \ and\ \ \ }
	\Def{E_\mu^\upsilon(2,\cdot)}{\upsilon(\cdot)}.
\end{gather}
The measure $\mu$ is that which we are interested in (we wish to compute
its ``entropy''); the measure $\upsilon$ is a choice we  make regarding
what to compare it against. Often $\upsilon=\lambda$, Lebesgue measure.
The \Def{unconstrained entropy} can be defined as follows.  For $D\in
\cvxrec(\R^2, \R-^2)$, the \Def{$D$-entropy of $\mu$ relative to
$\upsilon$} is 
\begin{gather}
	\Def{\ent_D^\upsilon(\mu)}{\I_D(E_\mu^\upsilon)}.
\end{gather}
Define $D_\phi$ via  \eqref{eq:D-phi-def} and
write $\Def{\ent^\upsilon_\phi(\mu)}{\ent_{D_\phi}^\upsilon(\mu)}$,
the
usual definition of $\phi$-entropy when $\upsilon$ is chosen to be
``uniform'' over the support of $\mu$.\footnote{This is not a new idea; see
	\citep[page 329]{chafai2004entropies}.} 
Choose $\rho$ as usual to be
absolutely continuous with respect to $\mu$ and $\upsilon$.  Then
using Proposition \ref{prop:dinf_support_representation} we have 
$ \ent_\phi^\upsilon(\mu) =
\int_X\sprt_{D_\phi}\left(\varfrac{\dv\mu}{\dv\rho}, 
\varfrac{\dv\upsilon}{\dv\rho}\right)\dv\rho                               
= \int_X \breve{\phi}\left(\varfrac{\dv\mu}{\dv\rho},
\varfrac{\dv\upsilon}{\dv\rho}\right)\dv\rho                               
= \int_X \varfrac{\dv\mu}{\dv\rho} 
    \phi\rbr{\varfrac{\dv\mu}{\dv\rho}\varfrac{\dv\rho}{\dv \upsilon}}\dv\rho
= \int_X\phi\rbr{\varfrac{\dv\mu}{\dv \upsilon}}\dv\mu           
=\E_\mu \phi\rbr{\varfrac{\dv\mu}{\dv \upsilon}}$.
As usual, the choice of reference measure $\rho$ does not matter.
But the choice of comparison measure $\upsilon$ does matter since
$\ent_\phi^\upsilon(\mu)=\I_\phi(E_\mu^\upsilon)$ which clearly
depends upon the choice of $\upsilon$. 

This perspective offers an insight into why
the entropy is difficult to estimate: one is implicitly attempting to
determine the Bayes risk for a statistical decision problem where the two
class conditional distributions are the given $\mu$ and the reference
(uniform) measure $\upsilon$ using a loss $\ell$ induced by $\phi$ as
in Remark \ref{rem:bridge-n-2}. This insight also offers an effective
approach to estimating the entropy as we now explain.

The \Def{constrained entropy} of $\mu$ relative to $\upsilon$ is defined similarly,
\begin{gather}
  \Def{\ent_{\scr F}^\upsilon(\mu)}{\I_{\scr F}(E_\mu^\upsilon)},
\end{gather}
and simply amounts to regularising the $\phi$-entropy (where $\scr
F(X)\subseteq D_\phi$). This immediately suggests ways to estimate the
entropy of a random variable defined on $X$ (especially when $X$ is high
dimensional): use the bridge between $\scr F$-information and the $\scr
H$-constrained Bayes risk and simply exploit the wide range of extant
methods for solving binary class-probability estimation problems.  That is
given a random sample $\{ x_1,\ldots,x_m \}$ drawn iid from $\mu$, estimate
the entropy from the empirical measure $\Def{\mu^m(A)}{\frac{1}{m}
\sum_{i\in[m]} \iver{x_i\in A}}$ via $\ent_{\scr F}^\upsilon(\mu^m)=\I_{\scr
F}(\mu^m,\upsilon)$.  The estimate is regularised by the choice of $\scr
F$.  Observe that one can immediately define a generalised mutual
information using $\I_{\mathcal{F}}$ when $n=2$: given two random variables
$\mathsf{Z}$ and $\mathsf{Y}$ defined on $X$  
with joint distribution $\mu_{\mathsf{ZY}}$
and marginal distributions $\mu_{\mathsf{Z}}$ and $\mu_{\mathsf{Y}}$,
define the experiment $E^{\mathrm{MI}}\colon[2]\expto X$ via 
$ \Def{E^{\mathrm{MI}}(1,\cdot) }{\mu_{\mathsf{ZY}}(\cdot)}$
and $\Def{E^{\mathrm{MI}}(2,\cdot) }{
(\mu_{\mathsf{Z}}\times\mu_{\mathsf{Y}})(\cdot)}$,
and then define  the \Def{$\mathcal{F}$-Mutual Information between
$\mathsf{Z}$ and $\mathsf{Y}$} as
\begin{gather}
\Def{\operatorname{MI}_{\mathcal{F}}(\mathsf{Z};\mathsf{Y})}{%
	\I_{\mathcal{F}}(E^{\mathrm{MI}})}.
\end{gather}
While this seems more complex then the usual notion of mutual information,
we observe that this is what is typically computed in practice since one
cannot ever find the Bayes optimal hypothesis implicit in the definition of
the usual mutual information, but rather only optimises over a restricted
model class.

Given that entropy can be reduced to binary divergences relative to an
arbitrarily chosen uniform measure, and further given the multitude of
binary divergences that make decision-theoretic sense, axiomatic arguments
for a single preferred  entropy are less compelling, nothwithstanding their
mathematical elegance \citep{Baez:2011aa}.

One can apply Proposition \ref{prop:I-F-ET} to $\scr F$-entropies where a
given distribution $\mu$ is pushed through a Markov kernel $T$ to give $\mu T$.
Since $\ent_{\scr F}^\upsilon(\mu)=\I_{\scr F}(E_\mu^\upsilon)$, 
we have $\I_{\scr F}(E_\mu^\upsilon T)= \I_{T^*\scr F}(E_\mu^\upsilon)$ and hence
\[
  {
      \color{ForestGreen}
      \ent_{\scr F}^\upsilon(\mu T) = \ent_{T^*\scr F}^\upsilon(\mu).
    }
\]

\section{Precursors of \texorpdfstring{$\mathcal F$}{F}-Information}
\label{sec:precursors}
There are several precursors\footnote{
	As we should well expect: 
	``far from being odd or curious or remarkable,
	the pattern of independent multiple discoveries in science is in
	principle the dominant pattern'' \citep[page 477]{Merton:1961aa}.
} to our notion of $\mathcal{F}$-information,
including {$\Nfrak$}-information (rediscovered as MMD), Integral 
Probability Metrics,
Moreau-Yosida {$\phi$}-divergences and $(f,\Gamma)$-Divergences, and in
this Appendix we briefly summarise them.

The idea that one can view a model class as being the result of a rich
class being ``pushed through'' a restrictive channel (what the
information processing equality does in effect) was central to the
calculations of covering numbers by \citet{Williamson:2001aa}.

As can be seen from \eqref{eq:classical-variational-form-no-abs} in
Appendix \ref{sec:var-rep}, the classical binary variational divergence
of $E\colon\cbr{1,2}\expto X$ can be written as
$
	\I_{\mathrm{var}}(E)=\sup_{f\colon X\rightarrow ([0,1],\borel)} E_1 f -
	E_2 f.
$
When the supremum is restricted to be over $\scr F$,  a proper subset of
$\{f\colon X\rightarrow ([0,1],\borel)\} $, these are known as
\Def{integral probability metrics} (IPMs) \citep{Muller:1997aa}
or \Def{probability metrics with $\zeta$-structure} \citep{Zolotarev:1983aa},
and extend the Variational divergence by restricting the class of functions
which are optimised over in its variational representation; 
see \ref{sec:variational-representation}. Special cases of
this include the Wasserstein distance \citep{Villani:2009aa}.

When $\scr F$ is the unit ball of a reproducing kernel Hilbert space, these
are known as \Def{$\Nfrak$-distances} and were developed by
\citet{Klebanov:2005aa, Baksajev:2010aa,Zinger:1992aa,Klebanov:2003aa};
\citep[see][Chapters 21--26, for a recent review]{Rachev:2013aa}. The
$\Nfrak$-distances were rediscovered in the machine learning community as
``Maximum Mean Discrepancy'' (MMD) by \citet{Smola:2007aa,
Sriperumbudur:2010aa, Vangeepuram:2010aa}. \citet{Muandet:2017aa} presented a
recent review (ignoring some prior work however).

The classical IPMs are a way of constraining the function class one
optimises over in the variational representation of variational divergence.
One can similarly restrict the class of functions in the variational
representation of an arbitrary $\phi$-divergence as was suggested by
\citet[page 796]{Reid2011},  who proposed considering 
$
	\Def{\I_{\phi,\mathcal{F}}(P,Q)}{\sup_{\rho\in\mathcal{F}} (\E_P\rho
-\E_Q \phi^*(\rho))},
$
explored the particular case for $\phi(t)=|t-1|$ and $\mathcal{F}$ being
the unit ball in a reproducing kernel Hilbert space \citep[Appendix
H]{Reid2011}, and posed the question of its relationship to a constrained
Bayes risk also using the function class $\mathcal{F}$ \citep[page
799]{Reid2011} (which is answered by the present paper).
\citet{Xu:2020th}  proposed a generalization of Shannon Mutual information by
restricting the class of functions optimised over in a variational
representation, motivated slightly differently to the
$\mathcal{F}$-information of the present paper --- they motivated their
definition on computational grounds, and observed as a consequence the
estimation performance improves. (Note the brief discussion of
$\mathcal{F}$-mutual information in Appendix \ref{sec:entropies}.)
\citet{Terjek:2021aa} regularised the
optimisation for binary $\phi$ divergences with a Wasserstein regulariser.
More generally, \citet{Birrell:2022} considered a larger range of
$\mathcal{F}$ for arbitrary $\phi$. However, they necessarily only
considered
the binary $\phi$-divergence, and because they used the classical
variational representation in terms of the Legendre-Fenchel conjugate of
$\phi$, their formulas become quite complex compared to the development in
the present paper.  A recent comparison of IPMs and $\phi$-divergence
\citep{Agrawal2021} appears to mix up two things: a comparison of loss
functions, combined with a question of the approximation power of a model
class.

The $\mathcal{F}$-information is also related to ideas developed in
distributionally robust optimisation, where existing divergences $d(P,Q)$
are ``smoothed.'' Three examples are the ``Gaussian smoothed sliced
Wasserstein distance'' of \citet{rakotomamonjy2021statistical} and the
``smoothed divergence'' defined either as $d^\epsilon(P,Q)=\max_{\rho\in
B_\epsilon(P)} d(\rho,Q)$ by \citet{Van-Der-Meer:2017te},  or (closest in
spirit to the present paper, especially the information processing
equality) $d^\epsilon(P,Q)=d(P*K_\epsilon, Q* K_\epsilon)$, where the $*$
denotes convolution and $K_\epsilon$ is a scaled kernel (of `width'
$\epsilon$) as defined by \citet[page 25]{manole2021sequential} 
\citep{goldfeld2020convergence} (motivated by effects of additive noise) and
\citet{nietert2021smooth}.  

The Wasserstein distance is  related to ``smoothed entropies''
\citep[equation 8]{Van-Der-Meer:2017te},  the idea of which is
to use  $d^\epsilon(P,Q) = \max_{P'\in B_\epsilon(P)} d(P',Q)$, where
in their case, $d$ is the Renyi divergence (related to, but different from
$\phi$-divergences), and the $\epsilon$-ball $B_\epsilon(P)$  is
relative to the trace distance. 


There are links between IPMs and distributional robustness motivated
by understanding $f$-GANs  \citep{Nowozin:2016aa}.  \citet{husain2019primal}
showed that ``restricted $f$-GAN objectives are lower bounds to Wasserstein
autoencoder.''  Subsequently \citet[Theorem 1]{husain2020distributional}
showed how the distributionally robust objective can be expressed via
regularisation:  $\sup_{Q\in B_{\epsilon,F}}\int h \dv Q = \int h \dv P +
\Lambda_{F,\epsilon}(h)$; see also \citep{song2020bridging}.


Finally we mention the perspective of \citet{Birrell:2022vm} who refined the
variational representation of $\phi$-divergences in a complementary way:
instead of restricting the function class over which the objective is
optimised, they tweak the form of the objective function in a manner that
the argmax remains the same, but the objective differs otherwise.

 \printbibliography
\end{document}